\documentclass[journal]{IEEEtran}
\usepackage{amsmath,amssymb,graphicx}
\usepackage{balance}
\usepackage{subfigure}
\usepackage{booktabs}
\usepackage{multirow}
\usepackage{color}
\usepackage{algorithm}
\usepackage{algpseudocode}
\ifCLASSINFOpdf
\else
\fi
\begin{document}
%
\title{Arc-support Line Segments Revisited: An Efficient High-quality Ellipse Detection}
%
%
%

\author{
Changsheng~Lu, 
Siyu~Xia,~\IEEEmembership{Member,~IEEE},
Ming Shao,~\IEEEmembership{Member,~IEEE}, and
Yun Fu,~\IEEEmembership{Fellow,~IEEE} %
\IEEEcompsocitemizethanks{
\IEEEcompsocthanksitem C. Lu is with the Key Laboratory of System Control and Information Processing, Ministry of Education of China, Shanghai Jiao Tong University, Shanghai 200240, China. E-mail: ChangshengLuu@gmail.com. 
\IEEEcompsocthanksitem S. Xia is with the School of Automation, Southeast University, Nanjing 210096, China. E-mail: xia081@gmail.com.
\IEEEcompsocthanksitem M. Shao is with the Department of Computer and Information Science, University of Massachusetts, Dartmouth, MA 02747, USA. Email: mshao@umassd.edu.
\IEEEcompsocthanksitem Y. Fu is with the Department of ECE, Northeastern University, Boston, MA 02115, USA. Email: yunfu@ece.neu.edu.}
\thanks{This work is supported by the National Natural Science Foundation of China under Grant 61671151 and 61728103.}
}

%
%

\markboth{IEEE TRANSACTIONS ON IMAGE PROCESSING}%
{Shell \MakeLowercase{\textit{et al.}}: Bare Demo of IEEEtran.cls for IEEE Journals}
%



\maketitle

\begin{abstract}
Over the years many ellipse detection algorithms spring up and are studied broadly, while the critical issue of detecting ellipses accurately and efficiently in real-world images remains a challenge. In this paper, we propose a valuable industry-oriented ellipse detector by arc-support line segments, which simultaneously reaches high detection accuracy and efficiency. To simplify the complicated curves in an image while retaining the general properties including convexity and polarity, the arc-support line segments are extracted, which grounds the successful detection of ellipses. The arc-support groups are formed by iteratively and robustly linking the arc-support line segments that latently belong to a common ellipse. Afterward, two complementary approaches, namely, locally selecting the arc-support group with higher saliency and globally searching all the valid paired groups, are adopted to fit the initial ellipses in a fast way. Then, the ellipse candidate set can be formulated by hierarchical clustering of 5D parameter space of initial ellipses. Finally, the salient ellipse candidates are selected and refined as detections subject to the stringent and effective verification. Extensive experiments on three public datasets are implemented and our method achieves the best F-measure scores compared to the state-of-the-art methods. The source code is available at https://github.com/AlanLuSun/High-quality-ellipse-detection.
\end{abstract}


%
\IEEEpeerreviewmaketitle

\section{Introduction}\label{sec:introduction}
%
%
%
%
\IEEEPARstart{E}{llipse} detection is a fundamental technique in image processing field and plays an indispensable role in shape detection and geometric measurement.
Actually, ellipse detector can be utilized to handle various real-world problems.
In PCB inspection field, one basic function of the defect detection machine is to precisely as well as fast locate the circular pads or holes. Moreover, accurate measurement of circular control points and elliptic fiducial markers is helpful to homography estimation and camera calibration~\cite{huang2016homography,calvet2016detection,heikkila2000geometric}, and some irregular objects could be fitted as ellipses to simplify the shape structure for efficient mathematical modeling \cite{crocco2016structure,da2010fitting,bai2009splitting,kothari2009automated}.
However, to our best knowledge, there exist few robust, stable, efficient and accurate ellipse detector algorithms to universally handle the ellipse detection problem in real-world images, which may have the presence of cluttered edges, motion blur, illumination, occlusion, noise and so on. The major reason is that an ellipse involves five parameters rather than that a circle just needs three, which results in detecting ellipse both efficiently and accurately to be a tough problem. Recently, convolutional neural network (CNN) \cite{lecun2015deep} is revolutionizing objection detection field, like Mask R-CNN \cite{he2017mask} and YOLO \cite{redmon2018yolov3}. Deep learning based methods can provide image proposals which contain oval objects for image pre-processing while they are still inappropriate to directly handle ellipse detection due to the issues of limited segmentation accuracy and expensive manual annotation. In most real-world applications, the practical requirements with regard to higher location accuracy and faster speed make ellipse detection problem even challenging.\par

\begin{figure*}[!t]
    \centering
    \includegraphics[width=\hsize]{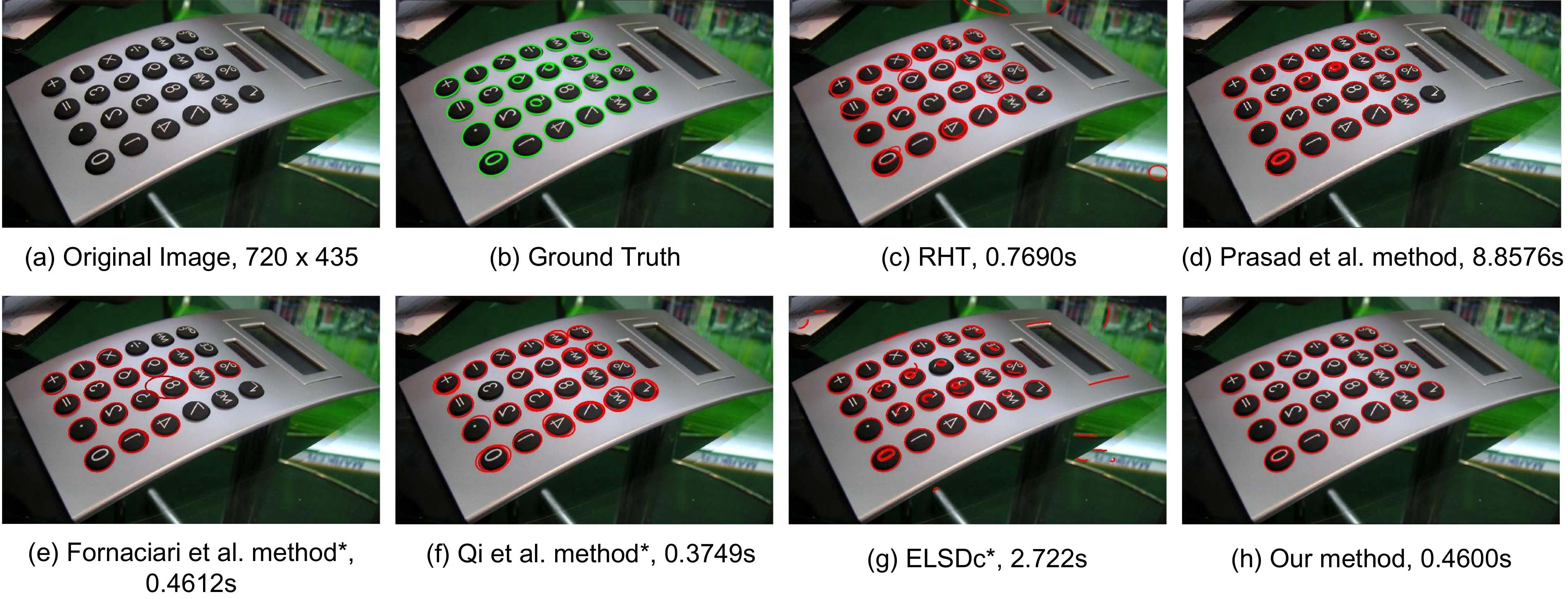}
    \caption{A comparison of various ellipse detection methods\protect\footnotemark[1]. The processing time of various methods is counted on the same computer with Intel Core i7-7500U 2.7GHz CPU and 8 GB memory. Except that the methods marked with (*) are implemented in C++, the remaining methods are in MATLAB. (a) the origin image is with the resolution of $720 \times 435$; (b) shows the ground true ellipses; (c) RHT \cite{mclaughlin1998randomized} can detect the ellipses while easily generating duplicates; (d) most ellipses can be located by Prasad et al. method \cite{prasad2012edge} at the cost of long running time; (e) and (f) show that the methods proposed by Fornaciari et al. \cite{fornaciari2014fast} and Qi et al. \cite{jia2017fast} are very fast. However, both methods cause either missing detections or false positives; (g) ELSDc \cite{patraucean2017joint} can jointly detect ellipses, arcs, and line segments while suffering from a long time; (h) our proposed method could accurately detect the ellipses with competitive running time, which reveals its high-quality detection performance.}
    \label{fig1}
\end{figure*}

Existing commonly used methods for ellipse detection can
be briefly grouped into two categories: 1) \text{\it{Hough Transform}}; 2) \text{\it{Edge Following}}.\par

Hough Transform (HT) has been widely used for detecting geometric primitives such as line segment (or LS for short), circle and ellipse~\cite{Duda1972Use}. The basic idea of HT for ellipse detection is voting arbitrary edge pixels into 5D parameter space. The local peak will occur when the corresponding bin of accumulator exceeds a threshold of votes, which implies for detecting an ellipse. But it is almost impractical to directly apply HT in practice due to the heavy computation burden and excessive consumption of memory. To alleviate these issues, considerable improved methods are put forward. Probabilistic Hough Transform (PHT) randomly selects a small subset of the edge points which is used as input for HT~\cite{kiryati1991probabilistic}, but large-scale attempts are taken to find the points all sharing a common ellipse and it leads to inferior performance when substantial noise exists. Yuen et al. decomposed the 5D parameter space by finding the ellipse center using some geometric properties like colinearity and symmetry on the first stage and then finding the remaining three parameters on the second stage~\cite{yuen1989detecting,tsuji1978detection}. Instead of transforming each edge point into a 5D parameter space, Xu et al.~\cite{xu1990RHT} proposed a Randomized Hough Transform (RHT) to detect curves, which randomly chooses five edge pixels each time and maps them into a point of the ellipse parameter space. McLaughlin et al. \cite{mclaughlin1998randomized} combined the aforementioned two-stage decomposition method and RHT at the aim of reducing the computation time and improving the detection performance compared with the standard HT, which becomes a baseline of ellipse detection method in the literature (Fig.~\ref{fig1}(c)). However, it is still not efficient enough in practice and always generates false detections due to the lack of novel validation strategies. Despite the simplicity of HT, HT based ellipse detection methods suffer from the following legacy problems: First, it is vulnerable in front of substantial image noise and complicated real-world background due to false peaks; Second, it requires much effort to tune the model parameters, e.g. bin size and peak threshold.\par
\footnotetext[1]{best viewed in color\label{footnote1}}

The second well-known family of ellipse detection methods is edge following, in which the connectivity between edge pixels, convexity of arc segments and geometric constraints are used. The general idea of edge following always starts from computing the \text{\it{binary edge}} map and corresponding gradients acquired by Canny or Sobel detector \cite{canny1986computational, plataniotis2013color} and then refining the arc segments from the \text{\it{binary edge}} for the ellipse fitting.\par

Many of edge following methods use line segments (LSs), which are extracted from the binary edges, as an intermediary to find the arc segments. The approach proposed by Kim et al. \cite{kim2002fast} merges the very short LSs to represent arc segments, where the arc fitting algorithms are frequently called. \cite{mai2008hierarchical} shares similar ideas with \cite{kim2002fast} to extract short LSs from the edge map while the difference lies in linking the LSs as well as the LS's edge points to form arc segments by using simple preset adjacency threshold and proper curvature condition. This method further iteratively groups two arc segments and applies the Random Sample Consensus (RANSAC) to the arc segment groups to recover the ellipse models. Although \cite{mai2008hierarchical} tries to promote the ellipse detector's robustness by iterative grouping and RANSAC, the massive missing detections (FNs) and false positives (FPs) appear.
The method proposed by Chia et al.~\cite{chia2011split} improves the framework illustrated in~\cite{mai2008hierarchical}, but a more complicated fragments merging and grouping procedures were employed. The merging of arc fragments is formalized as an alignment problem, where an alignment function is defined to score the rationality of merging, and a total cost function is built to incrementally search the optimal paired arc segments for grouping. Though the complex and iterative mathematical optimization boosts the detection performance to some extent, \cite{chia2011split} shows slow speed in the real-world images as reported in \cite{prasad2012edge,patraucean2017joint}. The ellipse detector proposed by Prasad et al. \cite{prasad2012edge} incorporates the edge curvature and convexity to extract smooth edge contours and performs a 2D HT to rank the edge contours in a group by the relationship scores for the better generation of ellipse hypotheses. But it also suffers from a long computation time, as shown in Fig.~\ref{fig1}(d).\par

Another stream of edge following methods tries to extract arc segments from binary edge directly and prunes straight edges for the purpose of fast detection speed. The ellipse detector proposed by Fornaciari et al. \cite{fornaciari2014fast} assigns a bounding box for each arc, removes the straight edges and determines the convexity of the arc by comparing the areas of region under and over the arc. In addition, this method accelerates the detection process by utilizing the property of that ellipse center should be colinear to the midpoints of parallel chords. However, it raises the detection speed at the cost of localization accuracy and robustness (Fig.~\ref{fig1}(e)). Recently, the method presented by Qi et al. \cite{jia2017fast} inherits \cite{fornaciari2014fast} and uses the similar convexity classification approach, but the difference lies in that \cite{jia2017fast} filters straight edges efficiently by calculating the edge connected component's characteristic number, which is a kind of projective invariant being able to distinguish the lines and conic curves within images. \cite{jia2017fast} is fast and yet prone to generate duplicates due to the absence of novel clustering (Fig.~\ref{fig1}(f)). In addition, both \cite{fornaciari2014fast} and \cite{jia2017fast} require at least three arc segments to recover the ellipse model, which might disable the algorithms when handling the incomplete ellipses.\par

Some researchers generalize the LS detection method to be a multi-functional detector which can jointly detect the LS and elliptic arcs. ELSDc proposed by P{\u{a}}tr{\u{a}}ucean et al. \cite{patraucean2017joint} uses an improved LSD \cite{grompone2010lsd} version for detecting LS, and then iteratively searches the remaining LSs from the start and end points of the detected LS. Eventually, both LS detection and grouping tasks are established simultaneously. Notably, ELSDc stands out other methods by detecting LSs from the \text{\it{greyscale image}} instead of \text{\it{binary edge}} such that abundant gradient and geometric cues can be fully exploited.
ELSDc and our proposed method are both based on LSD \cite{grompone2010lsd} for LS detection from the \text{\it{greyscale image}}, but they are fundamentally different from the generated LS type, ellipse candidates generation and validation strategies. Our method merely generates the arc-support LSs and do not chain them in the LS generation step. Moreover, ELSDc fits and validates the locally grouped LSs, omitting the global situation, which may be prone to produce the false positives (Fig.~\ref{fig1}(g)).\par

Arc-support LS is our previous work as introduced in \cite{Lu2017Circle}, each pair of which is successfully used for circle detection. However, it cannot handle the ellipse detection scheme since an arbitrary ellipse cannot be determined by two paired LSs. Admittedly, ellipse detection owns much higher complexity and requires more geometric cues. For example, the continuity feature, which is neglected in \cite{Lu2017Circle}, can be fully embodied in the arc-support group and is important in ellipse detection. Therefore, the careful arc-support groups forming, complicated geometric constraints, accurate ellipse generation and clustering, and novel validation strategy accustomed to ellipse detection are required, which will be addressed in this paper.\par

The main research purpose of this paper is to propose a high-quality ellipse detection method to handle the long-standing issue that cannot detect ellipses both accurately and efficiently in ellipse detection field. To that end, for the first time, we take the advantage of arc-support LSs for ellipse detection. The arc-support groups are formed by robustly linking the consecutive arc-support LSs which share similar geometric properties in point statistics level. Each arc-support group will be measured and assigned a saliency score. Secondly, we generate the initial ellipse set by two complementary approaches both locally and globally. The superposition principle of ellipse fitting and the novel geometric constraints, which are polarity constraint, region restriction and adaptive inliers criterion, are employed to consolidate the proposed method's accuracy and efficiency. Thirdly, we decompose the 5D ellipse parameter space into three subspaces according to ellipse center, orientation and semi-axes and perform three-stage efficient clustering. Finally, the candidates which pass the rigorous and effective verification will be refined by fitting again.\par

The rest of this paper is organized as follows. Section \ref{sec:Preliminary} introduces the preliminaries about arc-support LS and superposition principle of ellipse fitting. Section \ref{sec:ellipse detection} presents the high-quality ellipse detection framework, as a four-stage detection procedure: arc-support groups forming, initial ellipse set generation, clustering, and candidate verification. Section \ref{sec:complexity} analyzes the computation complexity of the proposed ellipse detection algorithm. Experimental results, as well as the accuracy and efficiency detection performance of the proposed method, are detailed in Section \ref{sec:experimental results}. Section \ref{sec:conclusion} concludes the paper.

\section{Preliminary}\label{sec:Preliminary}
In this section, the arc-support LS and its appendant properties are introduced as the basic geometric primitives for ellipse detection. Then we develop a superposition principle of fast ellipse fitting, which will save running time for the ellipse generation.
\subsection{Arc-support LS}
\begin{figure}[!t]
  \centering
  \includegraphics[width=\hsize]{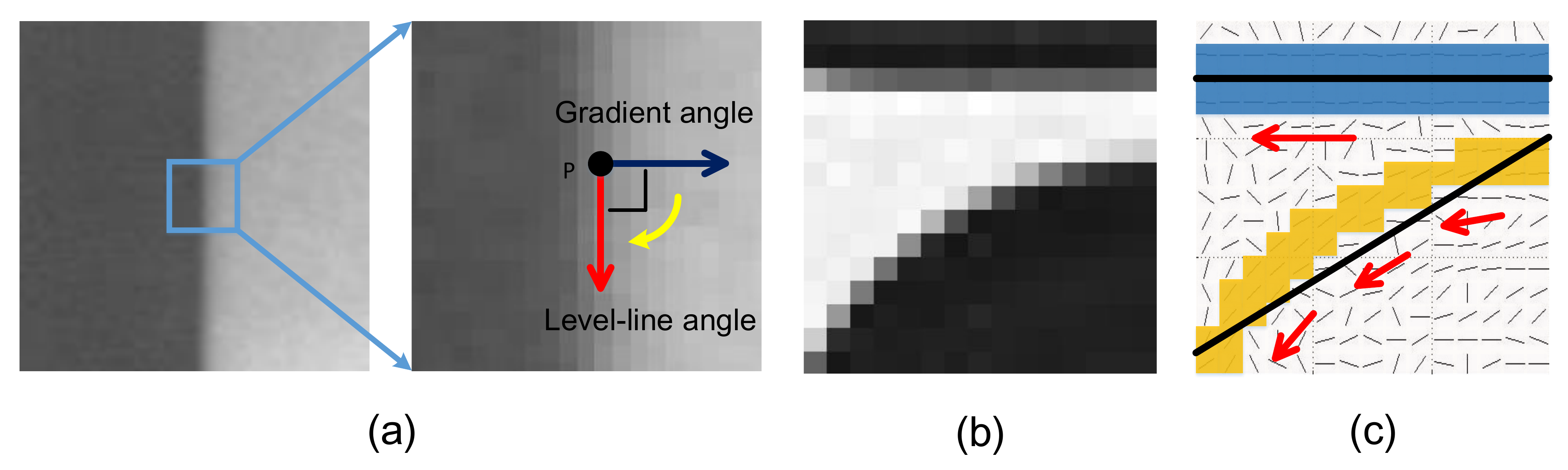}
  \caption{Level-line angle and two types of LS. (a) the level-line angle is acquired by clockwise rotating the gradient angle $90^{\circ}$; (b) greyscale image; (c) straight and arc-support LSs generated from (b).}
  \label{fig2}
\end{figure}

\begin{figure}[!t]
\centering
\includegraphics[width=\hsize]{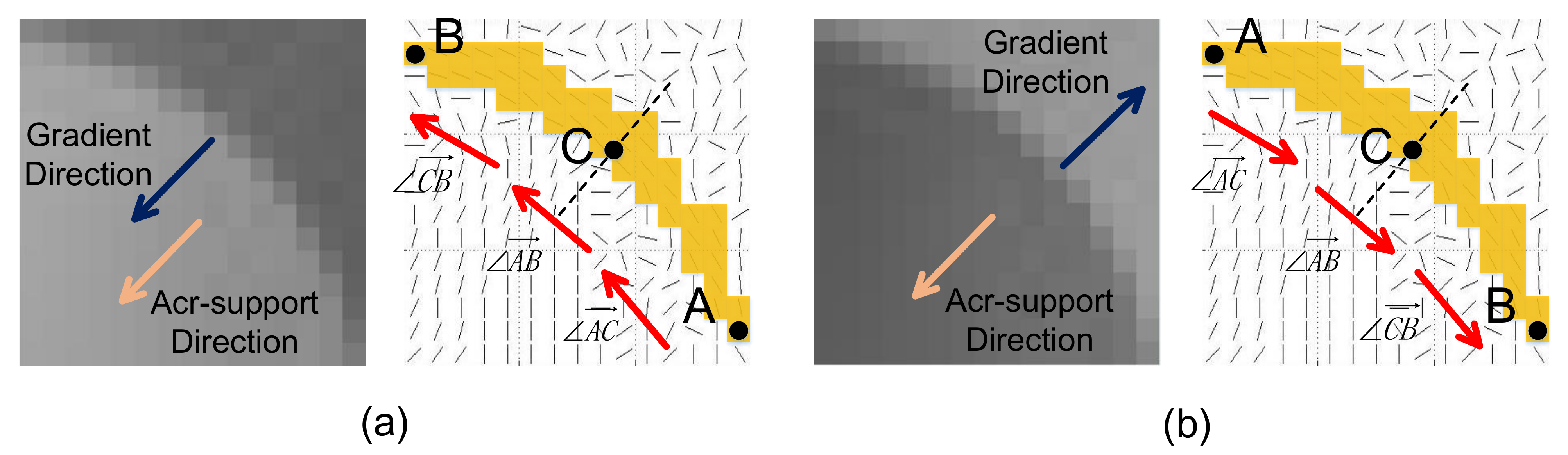}
\caption{Features of arc-support LS. (a) the overall gradient direction in the local greyscale area is same as arc-support direction and the three main angles in the corresponding level-line map change anticlockwise; (b) the conter-example of (a).}
\label{fig3}
\end{figure}

In image processing, LS mainly derives from two situations, as shown in Fig. \ref{fig2}. The first type LS comes from the support region where points share nearly the same level-line angle and overall distribute straight. Another type of LS derives from the arc-support region whose distribution changes like a curve. Thus, we call the LS approximated from arc-support region as ``arc-support LS''. Arc-support LS is built on top of LSD \cite{grompone2010lsd} as it is superior to other methods due to its efficiency and false control ability. The corresponding extraction procedures can be found in \cite{Lu2017Circle}. With the help of arc-support LS, the straight LS can be pruned while the arc geometric cues remain. Hereon, some properties of arc-support LS critical for ellipse detection are detailed.

\subsubsection{Arc-support Direction}
Different from conventional LS, arc-support LS carries the nature of convexity, standing for the ellipse center direction of an elliptic arc, namely the arc-support direction, as shown in Fig. \ref{fig3}. Assume that the two terminals of the circumscribed rectangle of the support region are $A$ and $B$ and the centroid is $C$. Thus the main angle of the support region is denoted as $\angle \overrightarrow{AB}$ and can be set to
\begin{equation}
\label{eq1}
\text{arctan}\left(\frac{{\sum\nolimits_{{p_i} \in \text{Region}} {\text{sin} (\text{level-line angle}({p_i}))} }}{{\sum\nolimits_{{p_i} \in \text{Region}} {\text{cos} (\text{level-line angle}({p_i}))} }}\right).
\end{equation}
Analogously, the main angles of two subregions $\angle \overrightarrow{AC}$ and $\angle \overrightarrow{CB}$ can be obtained according to Eq. (\ref{eq1}). Therefore, the arc-support direction can be set by  anticlockwise (or clockwise) rotating $\angle \overrightarrow{AB}$ by $90^{\circ}$ if $\angle \overrightarrow{AC}$, $\angle \overrightarrow{AB}$ and $\angle \overrightarrow{CB}$ change in the anticlockwise (or clockwise) direction and have an angle interval at least $T_{ai}$ in \{$\angle \overrightarrow{AC}$,$\angle \overrightarrow{AB}$ \} and \{$\angle \overrightarrow{AB}$,$\angle \overrightarrow{CB}$ \}.

\subsubsection{Polarity of Arc-support LS}
In the greyscale image, the overall gradient direction in the local area indicates the tendency of illumination variation. After the careful observation, there exist two situations between elliptic arc's overall gradient direction and arc-support direction. We define the polarity of an arc-support LS, namely $Pol_{L}$, is positive ($+1$) if the corresponding gradient direction and are-support direction are consistent, otherwise is negative ($-1$). A fast decision to the polarity of an arc-support LS is by judging the rotation direction of the main angles $\angle \overrightarrow{AC}$, $\angle \overrightarrow{AB}$ and $\angle \overrightarrow{CB}$, as shown in Fig. \ref{fig3}(a) and Fig. \ref{fig3}(b).

\subsection{Superposition Principle of Ellipse Fitting}
Ellipse fitting is very important in ellipse detection since it directly affects the quality of detected ellipse. Least-squares based ellipse fitting methods focus on minimizing the residue between points and ellipse \cite{rosin1993note,gander1994least,fitzgibbon1999direct}. As the constraint of ellipse fitting problem is quadratic, it usually leads to unsatisfactory efficiency along with the iterative procedure. Therefore, Fitzgibbon et al. \cite{fitzgibbon1999direct} proposed a non-iterative algorithm by solving the positive eigenvector of eigensystem. And we develop the superposition principle on the basis of \cite{fitzgibbon1999direct} due to its efficiency. Suppose that there are $n$ data points in the set $\Gamma_1  = \{ {p_1},{p_2}, \cdots ,{p_n}\}$, $p_{i}=\{ x_{i},y_{i} \}$. We first calculates $\Gamma_1$'s scatter matrix $\text{\bf{S}} =
\text{\bf{D}}^{\text{T}}\text{\bf{D}}$, and {\bf{D}} is denoted as
\begin{equation}
\label{eq2}
\text{\bf{D}} = {\left[ {\begin{array}{*{20}{c}}
{{x_1}^2}&{{x_1}{y_1}}&{{y_1}^2}&{{x_1}}&{{y_1}}&1\\
{{x_2}^2}&{{x_2}{y_2}}&{{y_2}^2}&{{x_2}}&{{y_2}}&1\\
 \vdots & \vdots & \vdots & \vdots & \vdots & \vdots \\
{{x_n}^2}&{{x_n}{y_n}}&{{y_n}^2}&{{x_n}}&{{y_n}}&1
\end{array}} \right]_{n\times6}}.
\end{equation}
Then by solving the generalized eigensystem $\text{\bf{S}}^{-1}\text{\bf{C}}$, where {\bf{C}} is the constant constraint matrix
\begin{equation}\label{eq:constant matrix}
\text{\bf{C}} = {\left[ {\begin{array}{*{20}{c}}
{\rm{0}}&{\rm{0}}&{{\rm{ - 1}}}& \cdots &{\rm{0}}\\
{\rm{0}}&{\rm{2}}&{\rm{0}}&{}&{}\\
{{\rm{ - 1}}}&{\rm{0}}&{\rm{0}}&{}& \vdots \\
 \vdots &{}&{}& \ddots &{}\\
{\rm{0}}&{}& \cdots &{}&{\rm{0}}
\end{array}} \right]_{{6\times6}}},
\end{equation}
the obtained eigenvector with positive eigenvalue is the desired fitted ellipse to $\Gamma_1$.\par

In practical ellipse detection process, it always needs to attempt fitting extensive different combinations of point sets for finding the most suitable fitted ellipse. Assuming $\Gamma_1$ has already been computed to fit an ellipse and after that several additional point sets belonging to the same ellipse are newly discovered, which are denoted by $\Gamma_2, \Gamma_3, \cdots, \Gamma_k$, an efficient computation approach to fit the new ellipse should be based on the previous computation results. Denote the design matrix and scatter matrix of $\Gamma_i$ as $\text{\bf{D}}(\Gamma_{i})$ and $\text{\bf{S}}(\Gamma_{i})$, respectively. Thus the design matrix $\text{\bf{D}}_{c}$ of the combination of $k$ point sets $\Gamma_1, \Gamma_2, \cdots, \Gamma_k$ can be written as
\begin{equation}
\label{eq3}
{\text{\bf{D}}_{c}} = \left[ {\begin{array}{*{20}{c}}
{\text{\bf{D}}({\Gamma_{1}})}\\
 \vdots \\
{\text{\bf{D}}({\Gamma_{k}})}
\end{array}} \right],
\end{equation}
and the corresponding scatter matrix $\text{\bf{S}}_{c}$ is
\begin{equation}
\begin{aligned}
\label{eq4}
{\text{\bf{S}}_c} &= \text{\bf{D}}_c^\text{T}{\text{\bf{D}}_c} = \text{\bf{D}}{({\Gamma_{1}})^\text{T}}\text{\bf D}({\Gamma_{1}}) +  \cdots  + \text{\bf D}{({\Gamma_{k}})^\text{T}}\text{\bf D}({\Gamma_{k}}) \\
      &= \text{\bf S}({\Gamma_{1}}) +  (\text{\bf S}({\Gamma_{2}}) + \cdots  + \text{\bf S}({\Gamma_{k}})).
\end{aligned}
\end{equation}
Eq. \ref{eq4} indicates that the scatter matrix of any combinatorial point sets equals the summation of the scatter matrix of each point set, which casts light on the feasibility of calculating the scatter matrix of each group merely once. The above superposition feature can cut computation time down when fitting one or more sets into an ellipse.

\section{High-quality Ellipse Detection}\label{sec:ellipse detection}
\begin{figure}[!tb]
\centering
\subfigure[]{
\includegraphics[width=0.28\hsize]{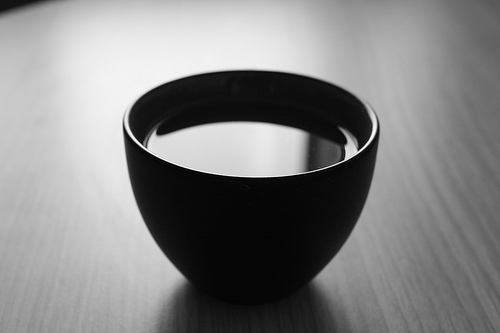}
\label{fig4:a}
}
\subfigure[]{
\includegraphics[width=0.28\hsize]{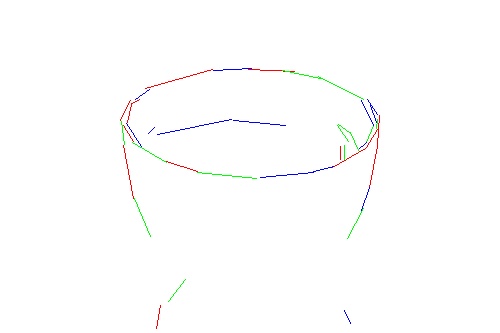}
\label{fig4:b}
}
\subfigure[]{
\includegraphics[width=0.28\hsize]{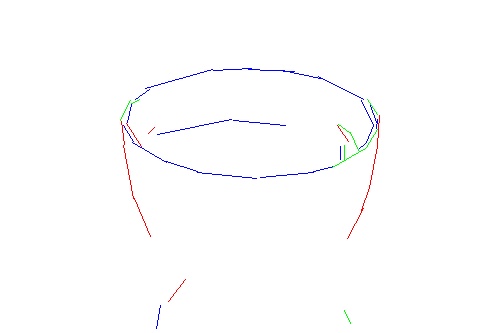}
\label{fig4:c}
}
\subfigure[]{
\includegraphics[width=0.28\hsize]{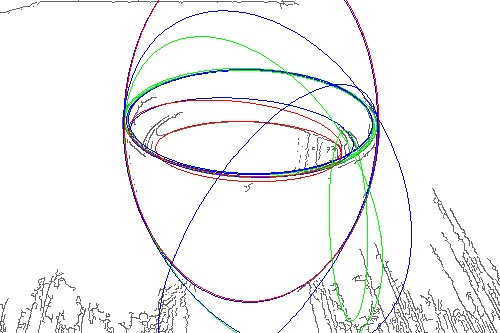}
\label{fig4:d}
}
\subfigure[]{
\includegraphics[width=0.28\hsize]{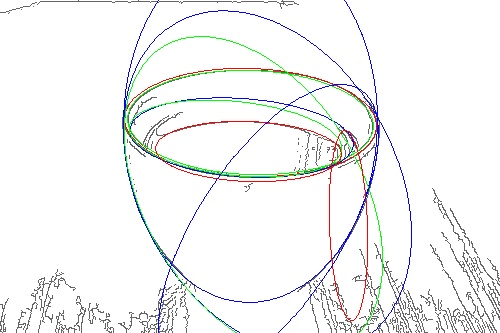}
\label{fig4:e}
}
\subfigure[]{
\includegraphics[width=0.28\hsize]{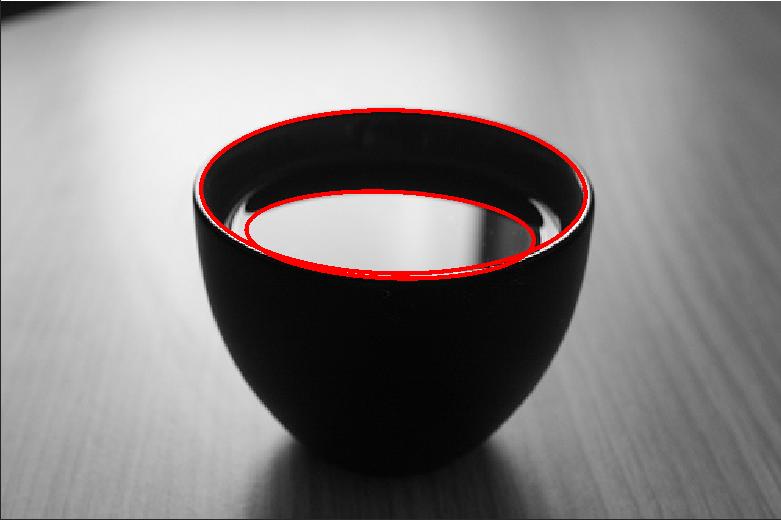}
\label{fig4:f}
}
\caption{Pipeline illustration of the proposed ellipse detection. (a) origin image; (b) 42 extracted arc-support LSs; (c) 20 arc-support groups; (d) 13 initial ellipses; (e) 10 ellipse candidates after clustering; (f) 2 qualified detections after verification and refinement.}
\label{fig4}
\end{figure}

In this section, a high-quality ellipse detection is proposed by introducing the arc-support LSs. The overall procedure consists of: (1) arc-support groups forming, (2) initial ellipse set generation, (3) clustering, and (4) candidate verification. The arc-support group collects the consecutive arc-support LSs belonging to the same curve, which can avoid the disturbance of the useless straight LSs. In the initial ellipse set generation step, accuracy and efficiency keep pace with the aid of fast ellipse fitting and effective geometric constraints. The efficient clustering and rigorous verification further facilitate the high detection performance of the proposed detector. An overall detection example of our method is demonstrated in Fig. \ref{fig4}.

\subsection{Arc-support Groups forming}
\subsubsection{Robust Linking and Groups Forming}
\begin{figure}[!bt]
\centering
\includegraphics[width=\hsize]{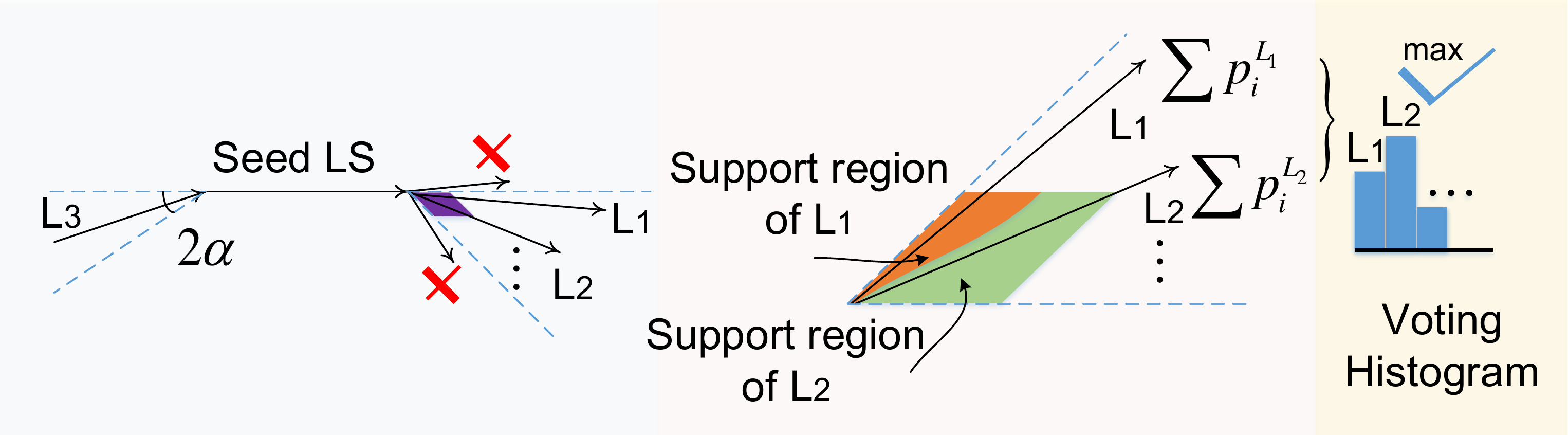}
\caption{The arc-support LSs are linked to form an arc-support group in point statistics level based on continuity and convexity.}
\label{fig5}
\end{figure}
Since an elliptic curve may consist of several arc-support LSs, we can link the discovered arc-support LSs to form a group. Any two consecutive arc-support LSs that will be linked should meet the continuity and convexity conditions. For continuity condition, the proximity between the head of an arc-support LS to the tail of another one should be close enough. For convexity condition, the linked LSs should change in the same direction either clockwise or anticlockwise.
Note that a support pixel's level-line angle should be within tolerance $\alpha$ with the support region's main angle, therefore, the angle deviation between two consecutive arc-support LSs should be less than $2\alpha$. To avoid incorrect LSs linking in the existence of noise, we count the number of support points of each next LS within a local statistical area near the terminal of current LS ($k$th LS's number of support points is represented as $\sum p_{i}^{L_{k}}$), and create a histogram for choosing the LS with maximum votes to link with current LS, as shown in Fig.~\ref{fig5}. Iteratively, the linked arc-support LSs which share the similar geometric properties are called as ``arc-support group''. Algorithm~\ref{algo1} details the arc-support LSs linking and groups forming process.\par
\begin{algorithm}[!b]
\caption{ Arc-support groups forming.}
\label{algo1}
\begin{algorithmic}[1]
\Require
Arc-support line segment set, $T_l$;
Arc-support regions that generate line segments, $T_r$;
Angle tolerance, $\alpha$;
Status where line segment used, $S$;
\Ensure
Arc-support groups, $G$;
\State Initialize groups $G = \varnothing$;
\label{code:alg1:initialization}
\Repeat
    \State Choose an arc-support line segment $l_i$ which satisfies $S(l_i) \neq used$ from $T_l$;
    \State Set the arc-support groups searched from the head and tail of $l_i$ as $g_{\text{head}}$ = $\varnothing$, $g_{\text{tail}}$ = $\varnothing$;
    \State Set $l_i$ as the seed of line segment $l_s$;
    \Repeat
    \State Searching consecutive arc-support line segments at the head end of $l_s$;
    \State Rule out the searched line segments which are $used$ and beyond $2\alpha$ angle deviation to $l_s$;
    \State Determine statistical area at the head end of $l_s$;
    \State Acquire the line segment $l_k$ with highest point votes by using $T_r$;
    \State Update $g_{\text{head}}$ = $g_{\text{head}}\cup L_{{k}}$, $S(l_k) = used$, $l_s$ = $l_k$;
    \Until{$l_s$ is $\varnothing$}
    \State Set $l_i$ as the seed of line segment $l_s$ again;
    \State $g_{\text{tail}}$ can be obtained by repeating the above searching process at the tail of $l_s$;
    \State Combine the searched arc-support groups $g_{\text{head}}$ = $\{L_{h1},\cdots,L_{hn}\}$ and $g_{\text{tail}}$ = $\{L_{t1},\cdots,L_{tn}\}$ as $g$ = $\{L_{tn},\cdots,L_{t1},L_{\text{i}},L_{h1},\cdots,L_{hn}\}$;
    \State Update $G$ = $G\cup g$;
    \State Update $S(l_i) = used$;
\Until{every arc-support line segment is traversed}\\
\Return $G$;
\end{algorithmic}
\end{algorithm}

\subsubsection{Spanning Angle Measurement for Each Group}
Each arc-support group which is composed of several arc-support LSs is essentially the polygonal approximation of a curve. If the $i$th group contains $n$ arc-support LSs, it will have $n-1$ angle intervals derived from every two consecutive arc-support LSs. Supposing that the angle interval sequence of $i$th group is $\{\theta^{i}_{1}, \theta^{i}_{2}, \cdots, \theta^{i}_{n-1}\}$, therefore, the spanning angle of $i$th group is $\sum\limits_{j = 1}^{n-1} {\theta _j^i}$. If an arc-support group is more salient to an ellipse, its spanning angle will be larger. Therefore, we have $S^{i}\propto (\sum\limits_{j = 1}^{n-1} {\theta _j^i})/{360^{\circ}}$,
where $S^{i}$ is the saliency score of the $i$th group. For convenience, the proportionality coefficient is set to $1$, and thus the range of $S^{i}$ is $[0,1]$.\par

As shown in Fig.~\ref{fig4:b}, only arc-support LSs are retained while the straight LSs are filtered. With the procedure of robust arc-support LSs linking, the LSs that belong to the same curves are pooled into arc-support groups (Fig.~\ref{fig4:c}).

\subsection{Initial Ellipse Set Generation}
Considering the fact that an arc-support group might contain all arc-support LSs of a curve, or merely a separate arc-support LS, therefore, we use two complementary methods to generate the initial ellipse set. First, from the local perspective, the arc-support LS group with relatively high saliency score is probably the dominant component of the polygonal approximation of an ellipse. A simple and effective manner is to individually fit the arc-support LS group to an ellipse completely relying on the threshold $T_{ss}$ such that the salient ellipse, for instance, the one with spanning angle close to $360^{\circ}$, can be picked out precedently. Second, from the global perspective, we search all the valid pairs of arc-support groups globally to reconstruct the latent ellipses on the image, one advantage of which is dealing with the troublesome situation of the arc-support groups for a common ellipse but far apart. All the valid pairs of groups should satisfy three criteria below: (1) polarity constraint; (2) region restriction; (3) adaptive inliers criterion. In addition, to avoid the overwhelming fitting process, the superposition principle is adopted during ellipse fitting for time efficiency.\par

\subsubsection{Polarity Constraint}
After observing the image regions around an elliptical edge, the inner of an ellipse is always either brighter or darker than the peripheral, where brighter means that the polarity of the arc-support LSs is positive and darker means the one is negative. If all arc-support LSs come from the same ellipse, generally, their polarity should be the same, too. Thus, the first criterion is that the polarity of the paired arc-support LS groups should be congruous. In certain cases, it may only require to detect the oval objects which are brighter (or darker) than background. Following our approach, the ellipse with specified polarity can be easily recognized, which will be detailed in Section \uppercase\expandafter{\romannumeral5}. E. As for the case of the arc-support groups derived from common ellipse while owning different polarity, e.g. an ellipse sharing different backgrounds, it still could be successfully detected as finding the most salient pair of arc-support groups with same polarity for ellipse fitting.\par

\subsubsection{Region Restriction}
Actually, most of the arc-support groups do not contribute to building a valid pair because of the high probability that two groups are from different ellipses or curves. Therefore, an early decision before fitting ellipse is essential.
\begin{figure}[!tb]
\centering
\includegraphics[width=0.8\hsize]{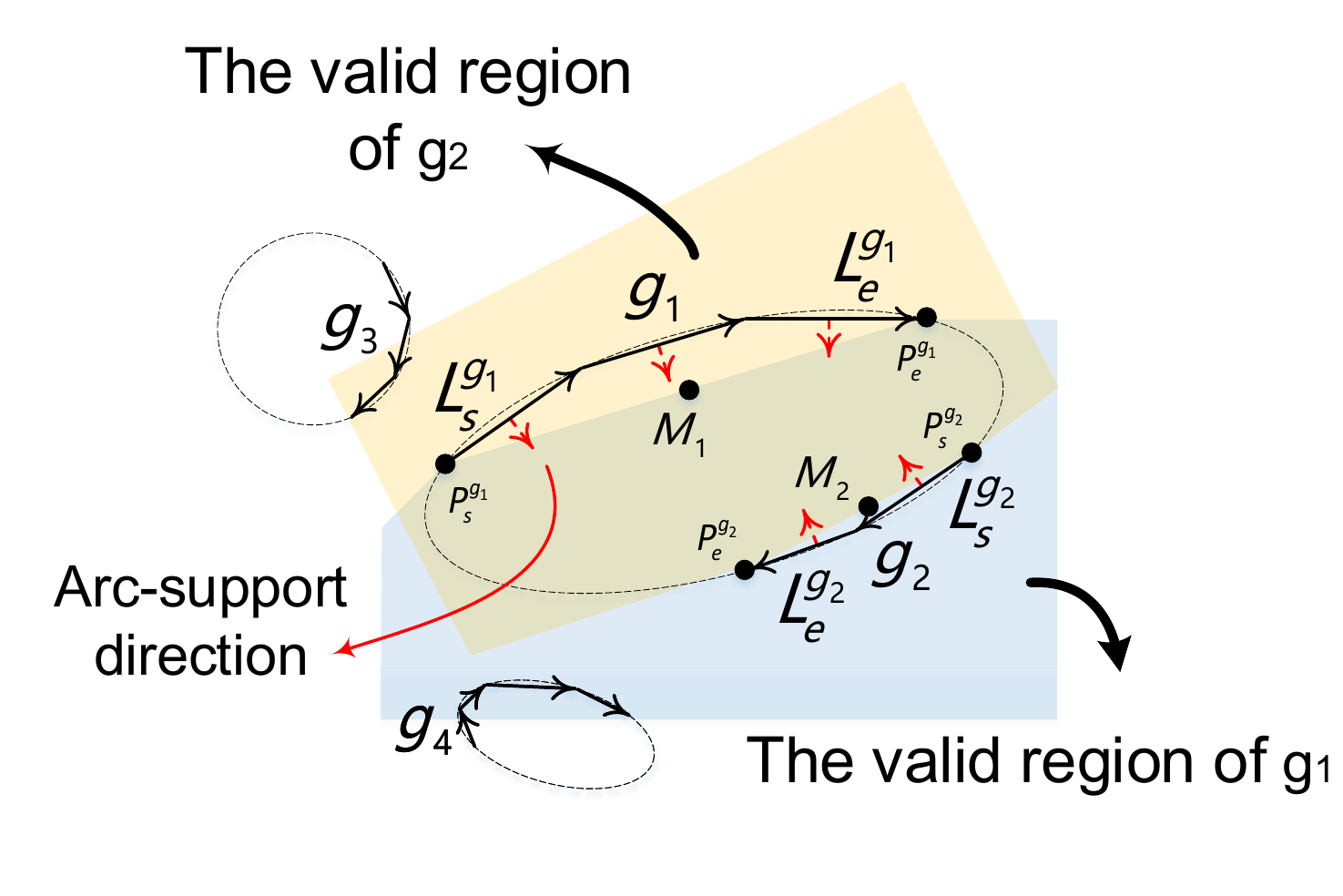}
\caption{Region restriction for a pair of two arc-support groups.}
\label{fig6}
\end{figure}
As shown in Fig.~\ref{fig6}, groups $\{g_{1},g_{3}\}$ and $\{g_{1},g_{4}\}$ should not be paired. The reason is that group $g_{3}$ is out of $g_{1}$'s valid region, which is along with the arc-support direction of every arc-support LS in $g_{1}$. Although $g_{4}$ is in the valid region of $g_{1}$, $g_{1}$ is not in the valid region of $g_{4}$. Consequently, if two groups are paired, they should locate in the mutual valid regions. In Fig.~\ref{fig6}, the start point of group $g_{i}$ is denoted as $P_{s}^{g_{i}}$ and the end point is $P_{e}^{g_{i}}$; the start arc-support LS is $L_{s}^{g_{i}}$ and the end arc-support LS is $L_{e}^{g_{i}}$; the midpoint of $P_{s}^{g_{i}}$ and $P_{e}^{g_{i}}$ is $M_{i}$. Then, we have
\begin{equation}
\label{eq6}
\begin{aligned}
  \left\{
  \begin{array}{c}
  \overrightarrow{ARC_{L_{s}^{g_{i}}}}\cdot\overrightarrow{P_{s}^{g_{i}}P_{e}^{g_{j}}} > \rho_{d} \\
  \overrightarrow{ARC_{L_{e}^{g_{i}}}}\cdot\overrightarrow{P_{e}^{g_{i}}P_{s}^{g_{j}}} > \rho_{d} \\
  (-Pol)\cdot Dir_{\bot}(\overrightarrow{P_{s}^{g_{i}}P_{e}^{g_{i}}})\cdot \overrightarrow{M_{i}M_{j}} > \rho_{d}
  \end{array}
  \right.
\end{aligned}
\end{equation}
where the $\overrightarrow{ARC_{L}}$ is the normalized arc-support direction vector of $L$. $Dir_{\bot}(\overrightarrow{v})$ represents the normalized vector acquired by rotating $\overrightarrow{v}$ clockwise $90^{\circ}$. The $Pol$ equals to $+1$ when the polarity of the pair $\{g_{1},g_{2}\}$ is positive, otherwise is $-1$. $\rho_{d}$ is distance threshold. Meanwhile, $(i,j)$ runs over $(1,2)$ and $(2,1)$.\par

\subsubsection{Adaptive Inliers Criterion}
For the pair $\{g_{1},g_{2}\}$ that passes the validations of polarity constraint and region restriction, an ellipse fitting against the endpoints of arc-support LSs in the pair will be implemented immediately. Assuming that the fitted ellipse is $e$, and the normal vector of edge point $P$ to $e$ is $\nabla e(P)$ (its direction points to the exterior of the ellipse),
we define edge point $P_{i}$ as a support inlier to $e$ if meeting both the distance tolerance $\epsilon$ and normal tolerance $\alpha$, namely the Rosin approximation distance \cite{rosin1998ellipse} from $P_{i}$ to $e$ should be less than $\epsilon$, and the absolute angle difference between $\nabla e(P_{i})$ and $-Pol_{e}\cdot Grad(P_{i})$ is less than $\alpha$, where $Pol_{e}$ is the polarity of $e$ and $Grad(P_{i})$ is image gradient of $P_{i}$. From the geometric perspective, we can approximate the arc length by the number of edge pixels. Therefore, the number of support inliers made up each arc-support LS in pair $\{g_{1},g_{2}\}$ should be greater than the corresponding LS's length, namely
\begin{equation}
\label{eq7}
\# \{p_i : p_i \in \text{SI}(L_{j})\} > \text{Length}(L_{j}),
\end{equation}
where $\text{SI}(L_{j})$ represents the support inliers set of arc-support LS $L_{j}$, $j = 1,2,\cdots,N_{g_{1}}+N_{g_{2}}$. $N_{g_{1}}$ and $N_{g_{2}}$ are the number of arc-support LSs in $g_{1}$ and $g_{2}$, respectively. In this step, the length of LS is an adaptive threshold for the corresponding support inliers. If an ellipse $e$ satisfies the adaptive inliers criterion, we will fit the support inliers to produce an initial ellipse. Eventually, all the valid pairs of arc-support groups are transformed to the initial ellipse set.

\subsection{Ellipse Clustering}
Considering the existence of duplicates in the initial ellipse set, an efficient clustering method is extremely important to trim them down, which should not only maintain the isolated points but also suppress the non-maximum. To that end, we develop a hierarchical clustering method based on mean shift~\cite{cheng1995mean}, which decomposes the 5D ellipse parameter space clustering problem into three low and cascaded dimensional space clustering problems (centers, orientations and semi-axes).\par

Assume that the initial ellipse set is $E^{\text{init}}$ and the element number of $E^{\text{init}}$ is $N^{\text{init}}$. We have
\begin{equation}
\label{eq8}
E^{\text{init}} = \bigcup_{1 \le i \le N^{\text{init}}}e_{i}
\end{equation}
where $e_{i} = \{(x,y)_{i}, \varphi_{i}, (a,b)_{i}\}$. Meanwhile, $(x,y)_{i}$, $\varphi_{i}$ and $(a,b)_{i}$ are the center, orientation and semi-axes of the initial ellipse $e_{i}$, respectively.\par
Firstly, our method clusters the ellipse centers of $E^{\text{init}}$ based on mean shift with limited iterations. It then produces $n^{\text{c}}$ elliptic cluster centers $(x,y)_{1}^{c}$, $(x,y)_{2}^{c}$, $\cdots$, $(x,y)_{n^{\text{c}}}^{c}$. If $e_{i}$ is the nearest to $(x,y)_{k}^{c}$, we will add $e_{i}$ to set $\Omega_{k}$. Therefore, $E^{\text{init}}$ can be divided into $n^{\text{c}}$ partitions. The $k$th ($1\le k \le n^{\text{c}}$) partition is represented as
\begin{equation}
\label{eq9}
\Omega_{k} = \{ e_{i} \parallel (x,y)_{i} \in e_{i} \; \text{and} \; (x,y)_{i}\rightarrow (x,y)_{k}^{c} \}.
\end{equation}
$(x,y)_{i} \rightarrow (x,y)_{k}^{c}$ means the Euclidean distance between data point $(x,y)_{i}$ and cluster center $(x,y)_{k}^{c}$ is least among all the cluster centers.\par

Secondly, each initial ellipse subset $\Omega_{k}$ is clustered with respect to their orientations. So $n^{\varphi}_{k}$ orientation cluster centers are generated, which are $\varphi^{c}_{1}, \varphi^{c}_{2}, \cdots, \varphi^{c}_{n^{\varphi}_{k}}$. In a similar way, $\Omega_{k}$ can be divided into $n^{\varphi}_{k}$ subsets. And the $s$th ($1\le s \le n^{\varphi}_{k}$) subset $\Omega_{k,s}$ corresponding to the orientation cluster center $\varphi^{c}_{s}$ is
\begin{equation}
\label{eq10}
\Omega_{k,s} = \{ e_{i} \parallel e_{i} \in \Omega_{k}, \varphi_{i} \in e_{i} \; \text{and} \; \varphi_{i}\rightarrow \varphi_{s}^{c} \}.
\end{equation}
\par

Finally, we implement the clustering step based on semi-axes for each initial ellipse subset $\Omega_{k,s}$ and therefore, $n^{\text{axes}}_{k,s}$ cluster centers are produced: $(a,b)^{c}_{1}$, $(a,b)^{c}_{2}$, $\cdots$, $(a,b)^{c}_{n^{\text{axes}}_{k,s}}$. Eventually, each mode of the combination $\{(x,y)^{c}_{k}, \varphi^{c}_{s}, (a,b)^{c}_{t}\}$ ($1\le t \le n^{\text{axes}}_{k,s}$) is the initial ellipse clustering result, namely the ellipse candidate. The ellipse candidate set $E^{c}$ can be described as
\begin{equation}
\label{eq11}
E^{c} = \bigcup_{k,s,t}\{ (x,y)^{c}_{k}, \varphi^{c}_{s}, (a,b)^{c}_{t} \}
\end{equation}
and the number of ellipse candidates $N^{c}$ is
\begin{equation}
\label{eq12}
{N^{c}} = \sum\limits_{k} {\sum\limits_{s}{n_{k,{\rm{ }}s}^{\text{axes}}} }.
\end{equation}
\par
As a result, the purer ellipse candidates can be generated after applying the hierarchical clustering approach to initial ellipse set. The benefits of such ellipse parameter space decomposition are two folds.
First, we distinguish the different geometrical significance of ellipse center, orientation, and semi-axes for hierarchical clustering since it is difficult to assign an accurate distance measure between two ellipses directly. Second, our clustering method is conveniently implemented and its computation complexity is quadratic (details can be found in Section \uppercase\expandafter{\romannumeral4}), which vastly outperforms than the direct clustering behavior in 5D space.

\subsection{Ellipse Candidate Verification}
The learned ellipse candidates cover almost all possible ellipses existing in the image while keeping nearly few duplicates, which saves much verification time and guarantees good recalls due to their pure quantity and high saliency. In this section, to further ensure the quality of the detected ellipses, we conduct the ellipse candidate verification that incorporates the stringent regulations for goodness measurement and elliptic geometric properties for refinement.

\subsubsection{Goodness Measurement}
The study has shown positive correlation between the number of support inliers on an ellipse and corresponding perimeter $\mathcal{B}$~\cite{kulpa1979properties}. Although the precise numerical formula of $\mathcal{B}$ has not yet been revealed, in practice, it can use $\mathcal{B} \approx \pi[\frac{3}{2}(a+b)-\sqrt{ab}]$ for approximation where $a$ and $b$ are the semi-major axis and semi-minor axis respectively \cite{zhang2016consistency}. In addition, the larger the angular coverage $\mathcal{C}$ of the elliptic connected component of support inliers becomes, the more salient the ellipse candidate will be. Based on these facts, we employ the following evaluation regarding both the number of support inliers and the angular coverage. The support inliers should satisfy the distance tolerance $\frac{\epsilon}{2}$ and the normal tolerance $\alpha$. Therefore, the ``goodness'' can be formulated as
\begin{equation}
\label{eq13}
\text{Goodness}(e)=\sqrt{\frac{\# \{p_i : p_i \in \text{SI}(e)\}}{\mathcal{B}} \cdot \frac{\mathcal{C}}{360^{\circ}}}
\end{equation}
where $\text{SI}(e)$ represents the support inliers of ellipse $e$. Note that we tighten the distance tolerance $\epsilon$ to $\frac{\epsilon}{2}$ to rule out the noise, and thus the goodness measurement is more credible. Finally, we apply a pseudo descending order (in linear time) to the ellipse candidates according to their goodness scores and preferentially pick out the candidates for subsequent validation.

\subsubsection{Verification and Fitting Again}
In verification, the proposed method continues to loose the distance tolerance $\frac{\epsilon}{2}$ to $\epsilon$, and validate the candidate individually against: (1) the number of support inliers; (2) the angular coverage of ellipse. We use a ratio threshold $T_{r}$ and expect that there are $T_{r}\mathcal{B}$ support inliers on the ellipse. In the meanwhile, our method only accepts the ellipse whose angular coverage is at least $T_{ac}$ degrees.\par

Given sufficient support inliers, we may have better ellipse results. Recall that we have found the support inliers with respect to each candidate in the validation step. If a candidate generates the true ellipse (or TP for short), its support inliers should be more sufficient than the previous one. This motivates us to fit ellipse again, which improves the overall accuracy and shows self-calibrated ability. Notably, the time complexity of the additional fitting only relates to the number of final detected ellipses, which will not bring on significant changes in the running time.

\section{Complexity of The Algorithm}\label{sec:complexity}
Assuming that the image size is $N \times N$, therefore, the arc-support LS extraction approach has an $O(N^{2})$ complexity equivalent to LSD \cite{grompone2010lsd}. In the procedure of the arc-support groups forming, the computational complexity is $O(N_{L})$, where $N_{L}$ denotes the number of arc-support LS. If there are $N_{G}$ groups, spanning angle measurement of groups has the complexity of $O(N_{G})$. In the initial ellipse set generation step, the computational complexity has upper bound $O(N_{G}^{2})$ in the worst case, which may rarely occur due to the three novel geometric constraints. In the ellipse clustering step, the running time has the complexity of $O(1+n^{\text{c}}+\sum \limits_{k}n^{\varphi}_{k})$. Noting that after clustering, the number of cluster centers is less than or equal to the number of original data points, which means that $n^{\text{c}} \leq N^{\text{init}}$ and $\sum \limits_{k}n^{\varphi}_{k} \leq N^{\text{init}}$. Note that the initial ellipses' number $N^{\text{init}}$ is of $O(N_{G}^{2})$, and thus the process of initial ellipse clustering step has the computational complexity of $O(2N_{G}^{2})$. In the validation step, the computation time relates to the number of candidates $N^{c}$. Eventually, the computational complexity of the proposed method is upper bounded by $O( N^{2} + N_{L} + 3N_{G}^{2} + N^{c})$, which reveals that the ellipse detection complexity is as fast as quadratic in $N$ and $N_{G}$.

\section{Experimental Results}\label{sec:experimental results}
In this section, extensive and detailed experiments are implemented to demonstrate the high-quality ellipse detection performance of the proposed method compared to the existing state-of-the-art methods.\par

\subsection{Experiments Setup}
\begin{table*}[!t]
  \centering
  \newcommand{\tabincell}[2]{\begin{tabular}{@{}#1@{}}#2\end{tabular}}
  \caption{Ellipse Detection Results on Three Public Real-world Datasets Compared with the Popular and State-of-the-art Methods.}
  \label{tab1}
  \begin{tabular}{cccccccc}
    \toprule[1pt]
    Dataset                             &Index     & RHT \cite{mclaughlin1998randomized}     & ELSDc \cite{patraucean2017joint}  & Prasad \cite{prasad2012edge}  & Fornaciari \cite{fornaciari2014fast}  & Qi \cite{jia2017fast}              & Our \\\midrule[1pt]
    \multirow{4}*{Traffic Sign Dataset} &Precision & 0.0706  & 0.0429 & 0.1401  & 0.4545      & 0.5814  & \textbf{\textbf{0.9110}} \\
                                        &Recall    & 0.5149  & 0.6933 & 0.5665  & 0.7277      & 0.7324  & \textbf{0.8811} \\
                                        &F-measure & 0.1242  & 0.0808 & 0.2246  & 0.5596      & 0.6482  & \textbf{0.8958} \\
                                        &Time/s     & 1.2875  & 4.6006 & 9.9918  & 0.2875      & \textbf{0.1488} & 0.5791 \\\midrule[0.5pt]
    \multirow{4}*{Prasad Dataset}       &Precision & 0.1941  & 0.0922 & 0.2360  & 0.7039      & 0.7161 & \textbf{0.7523} \\
                                        &Recall    & 0.2479  & 0.2940 & 0.3145  & 0.2154      & 0.2393 & \textbf{0.3504} \\
                                        &F-measure & 0.2177  & 0.1403 & 0.2697  & 0.3298      & 0.3587 & \textbf{0.4781} \\
                                        &Time/s     & 0.2011  & 1.4348 & 3.9011  & 0.0939      & \textbf{0.0529}  & 0.1685 \\\midrule[0.5pt]
    \multirow{4}*{\tabincell{c}{PCB Dataset}}&Precision&0.4165 &0.4168 &0.7645 &0.8933    & 0.9393        &\textbf{0.9715}\\
                                        &Recall    & \textbf{0.9500}  & 0.9154 & 0.7115  & 0.8692      & 0.8923           & 0.9192 \\
                                        &F-measure & 0.5791  & 0.5728 & 0.7371  & 0.8811      & 0.9152           & \textbf{0.9447} \\
                                        &Time/s      & 0.2271  & 3.4830 & 1.0758  & 0.0980      & \textbf{0.0738}           & 0.1693
    \\\bottomrule[1pt]
  \end{tabular}
\end{table*}
\subsubsection{Model Parameters}
Our ellipse detection method mainly involves seven parameters, which are discussed as follows:

(1) $T_{ai}$ is the angle interval of each subregion to the support region, which is used in arc-support LS extraction. It indicates the least curvature degree of the support region that generates an arc-support LS. Obviously, the LS derived from the straight edge will be filtered since its angle interval is less than $T_{ai}$. We fix $T_{ai}$ to $2.25^{\circ}$ as it performs well in experiments. (2) Angle tolerance $\alpha$ is used in the cases when evaluating the angle deviation of a geometric primitive, e.g. the point's level-line angle or gradient angle, to the corresponding reference angle. $\alpha$ can be empirically set to $22.5^\circ$ which yields the best results for thousands of images~\cite{grompone2010lsd,burns1986extracting}. (3) Saliency score threshold $T_{ss}$ is used in the initial ellipse set generation. Admittedly, we can fit any group as long as its saliency score is higher than zero because of the negligible computational cost. However, it is unnecessary because we will find all the valid pairs of arc-support groups for generating the initial ellipses. Therefore, we set $T_{ss}$ to $0.25$. (4) Distance threshold $\rho_{d}$ is used in region restriction, which can be minus for dealing with some extreme cases. Thus $\rho_{d}$ is fixed to $-3\epsilon$. (5) The distance tolerance $\epsilon$ is used to recover the inliers to arc-support LS or ellipse. If $\epsilon$ becomes more restrictive, the inliers will be purer, even in presence of spurious edges or noise. As our algorithm aims at high localization accuracy, we are able to set $\epsilon$ to $2$ pixels. (6) $T_{r}$ is the ratio of support inliers on an ellipse and (7) $T_{ac}$ is the elliptic angular coverage threshold. They both are used in the ellipse verification. $T_{r}$ and $T_{ac}$ are easily tuned due to their geometric significance. Since most of the true ellipses in real-world images have a degree of completeness, we set $T_{r} = 0.6$ and $T_{ac} = 165^{\circ}$ in default.\par

Notably, we merely open \text{\it{two}} external adjustable parameters $T_r$ and $T_{ac}$ after regarding the other five parameters as intrinsic parameters since they can be empirically fixed and work fairly well, which enables our ellipse detection algorithm to be easily used.

\subsubsection{Evaluation Metrics}
We employ the following metrics for evaluations: (1) \textbf{ precision}, (2) \textbf{recall}, (3) \textbf{F-measure}. The precision = TPs$/$(TPs + FPs), recall = TPs$/$(TPs + FNs) and F-measure = $2/$(precision${}^{-1}$ + recall${}^{-1}$). A detected ellipse is regarded as a TP if its overlap area ratio to the corresponding ground true ellipse is larger than $D_{0}$. And we set $D_{0}$ to $0.8$ throughout all experiments as did in \cite{prasad2012edge,fornaciari2014fast, jia2017fast}.

\subsubsection{Compared Methods}
We mainly select five most popular and competitive algorithms for quantitative and qualitative comparisons, which are RHT \cite{mclaughlin1998randomized}, ELSDc \cite{patraucean2017joint} and the methods proposed by Prasad et al. \cite{prasad2012edge}, Fornaciari et al. \cite{fornaciari2014fast}, and Qi et al. \cite{jia2017fast}. RHT is the most popular in the literature and often used as the baseline method. ELSDc is robust and achieves precise location accuracy amongst state-of-the-art methods. Prasad method combines HT and the techniques of edge following, which reflects very well detection performance. The methods proposed by Fornaciari et al. and Qi et al. are very efficient and obtain extremely high F-measure scores on public datasets. We compare our method\footnote[2]{The source code and more detection examples of our method can be found at https://github.com/AlanLuSun/High-quality-ellipse-detection.\label{footnote2}} to the aforementioned ellipse detectors as they are the most popular or state-of-the-art detectors existing in ellipse detection field.\par

For the fair comparison purpose, we adopt the source codes of ELSDc \cite{patraucean2017joint}, Prasad et al. \cite{prasad2012edge}, Fornaciari et al.~\cite{fornaciari2014fast}, and Qi et al. \cite{jia2017fast} which are available online, and reimplement RHT \cite{mclaughlin1998randomized} according to the original paper. RHT, Prasad method and our method are run in MATLAB while the remaining three methods are in C++. All experiments are performed with default parameters and on the same computer with Intel Core i7-7500U 2.7GHz CPU and 8 GB memory.

\subsubsection{Datasets}
To test the competitive ellipse detectors, three public challenging real-world datasets are utilized:
\begin{itemize}
  \item \textbf{Traffic sign dataset}. As a portion of Dataset $\sharp2$ which is created by Fornaciari et al. \cite{fornaciari2014fast}, it contains 273 images with various ellipses that are projected by round traffic signs at different real-life scenarios. These images derive from the frames of several videos captured by smartphone, 
      suffering from the blurry and varying lighting conditions by motion and autofocus.
  \item \textbf{Prasad dataset}. A complex real-world image dataset, employed before by many well-known ellipse detectors in their experiments~\cite{prasad2012edge,fornaciari2014fast, patraucean2017joint, jia2017fast}. Note that Prasad dataset consists of 198 images which are complex enough due to the unpredictable conditions and substantial disturbances.
  \item \textbf{PCB dataset}. PCB (Printed Circuit Board) dataset \cite{Lu2017Circle} includes 100 industrial PCB images with various disturbances and each image contains at least one circular or elliptic shape. All PCB images are labeled manually and precisely.
\end{itemize}
\par

\subsection{Experiments on Real-world Datasets}
\begin{figure*}[!tb]
\centering
\includegraphics[width=0.9\hsize]{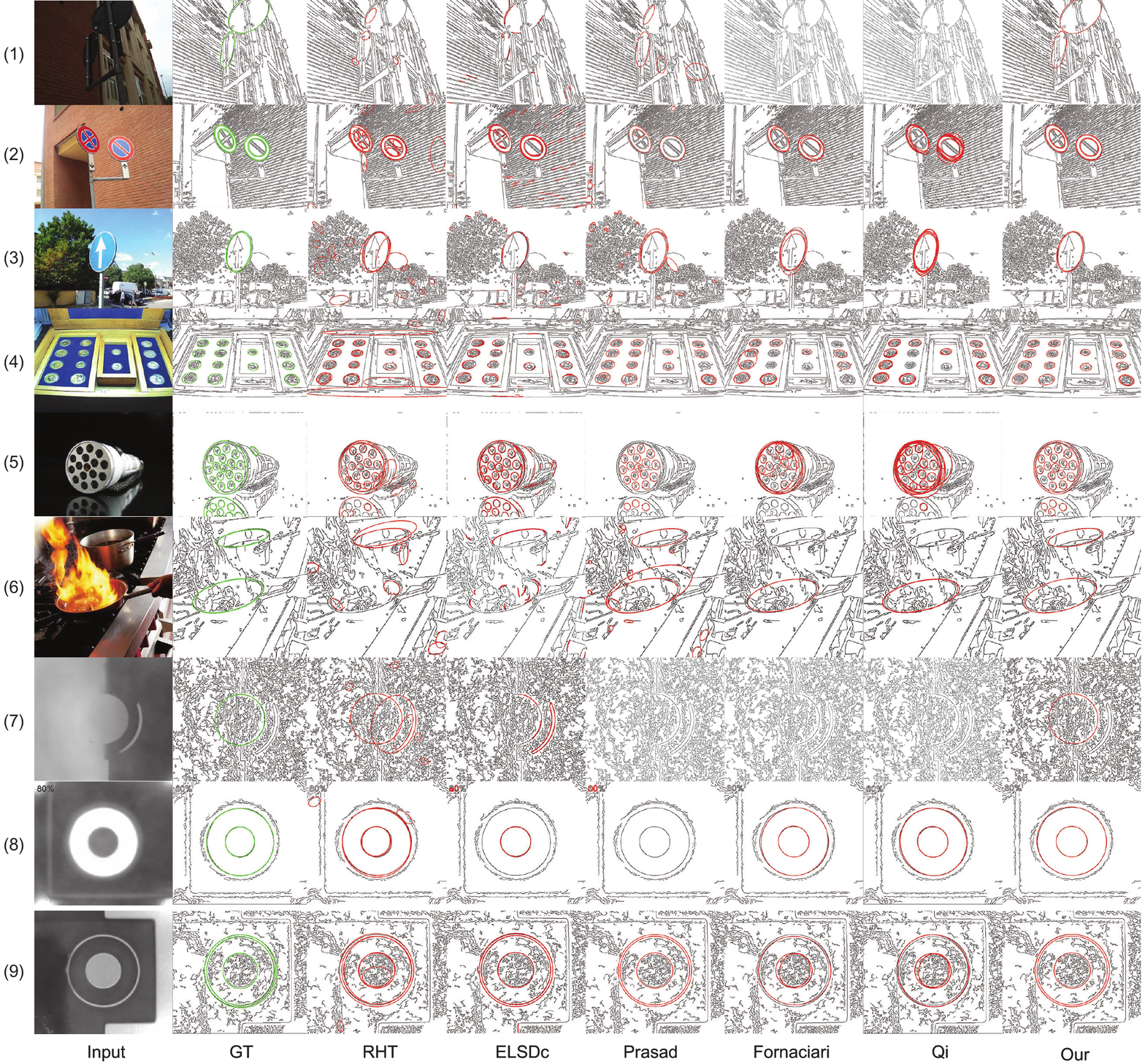}
\caption{Ellipse detection examples on three real-world datasets. The first and second columns are the input images and ground truth (GT). The input images of the first three rows are from traffic sign dataset while the second and third three-row images are from Prasad dataset and PCB dataset.}
\label{fig7}
\end{figure*}

\begin{figure*}[!tb]
\centering
\subfigure[]{\includegraphics[width=0.3\textwidth]{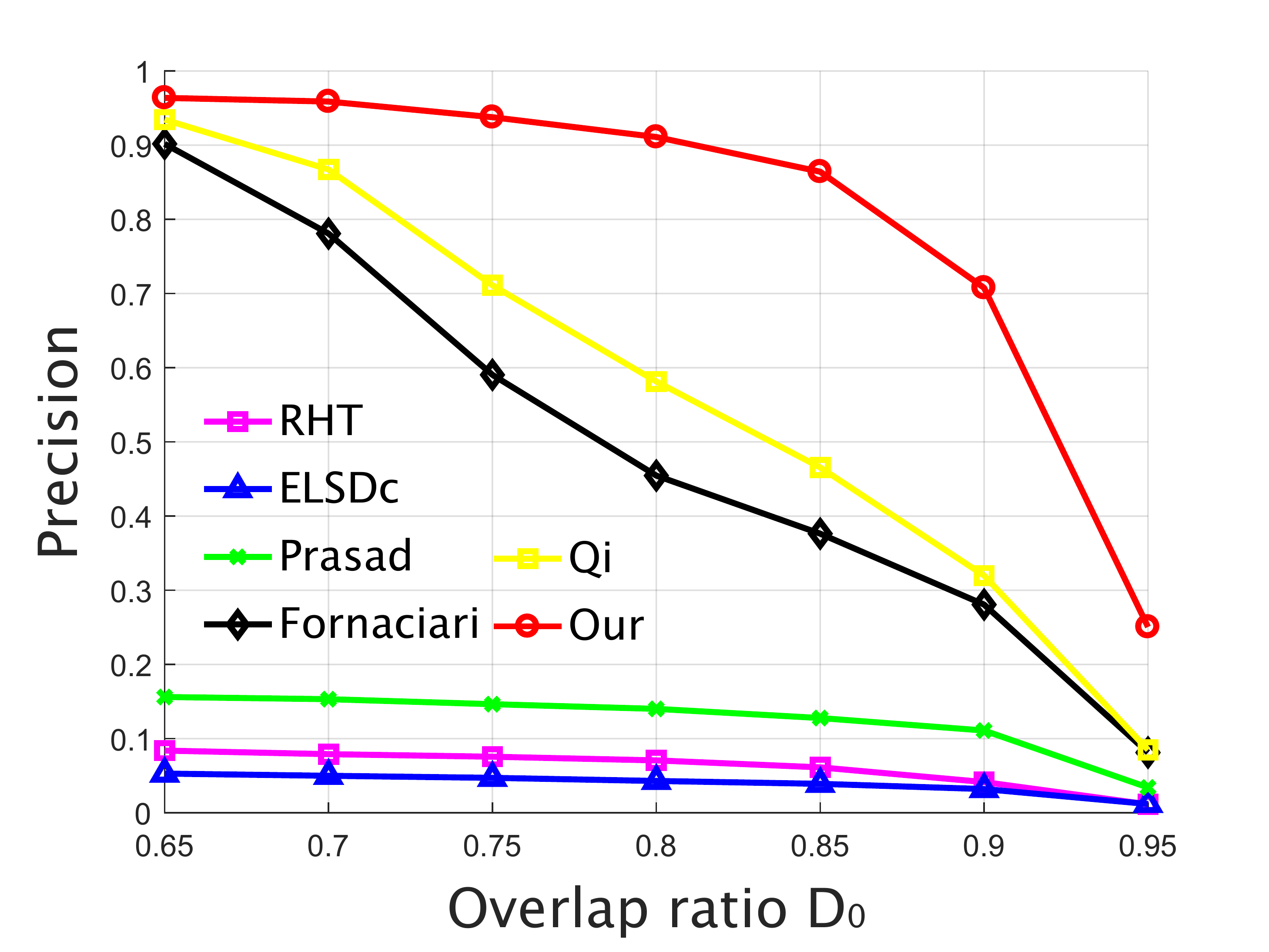}
\label{fig8:Traffic-P}}
\subfigure[]{\includegraphics[width=0.3\textwidth]{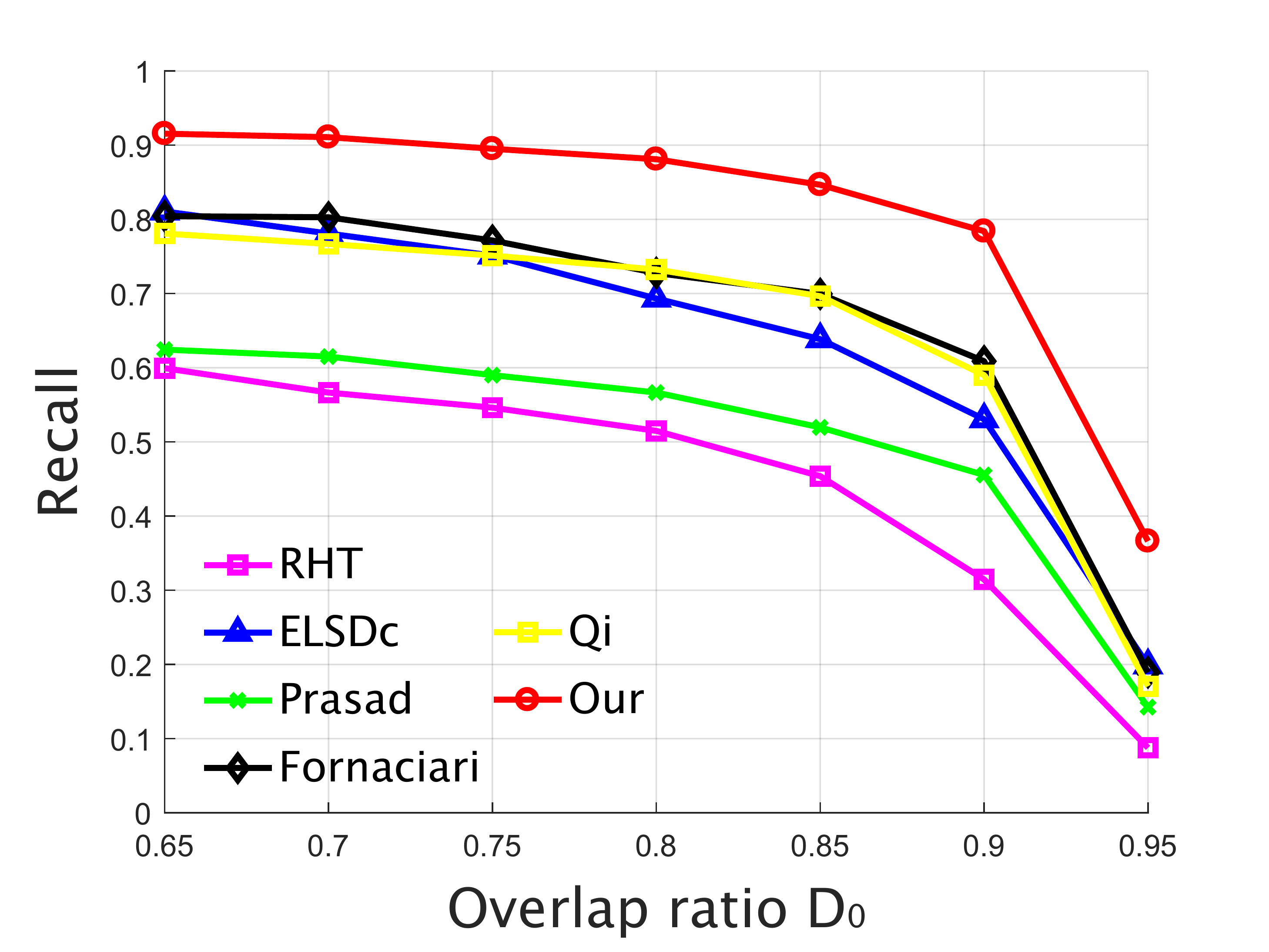}
\label{fig8:Traffic-R}}
\subfigure[]{\includegraphics[width=0.3\textwidth]{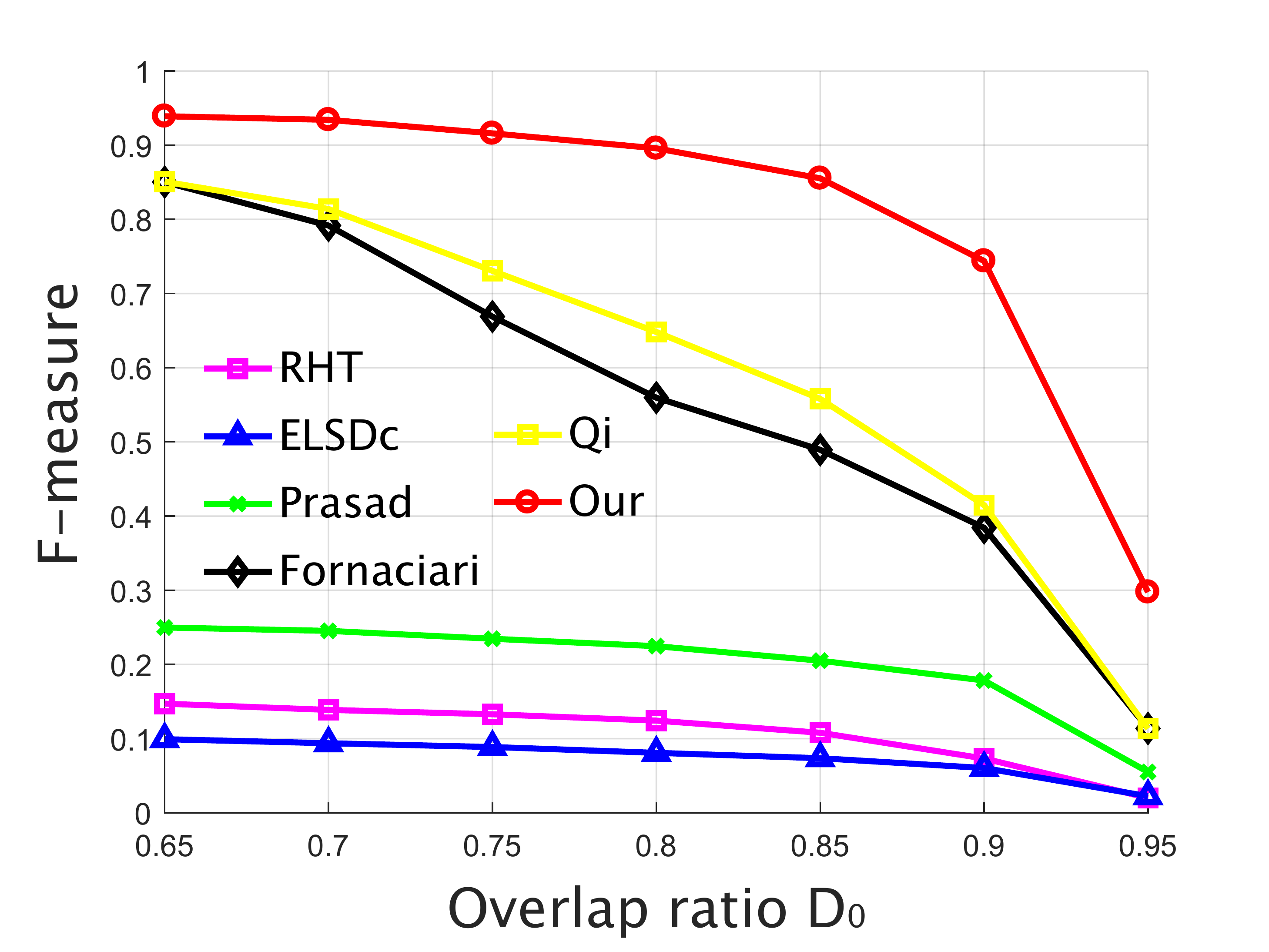}
\label{fig8:Traffic-F}}
\subfigure[]{\includegraphics[width=0.3\hsize]{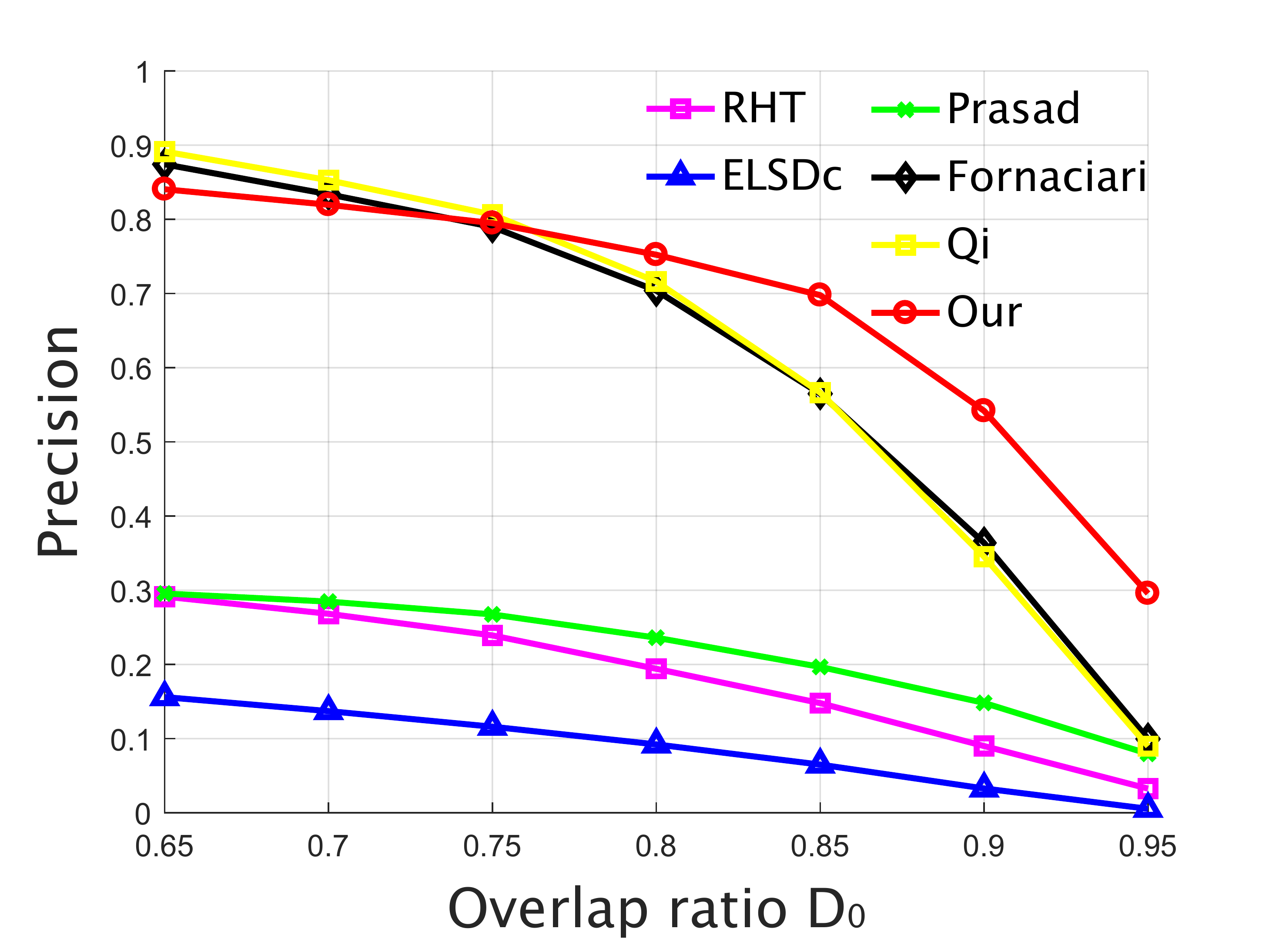}
\label{fig8:Prasad-P}}
\subfigure[]{\includegraphics[width=0.3\hsize]{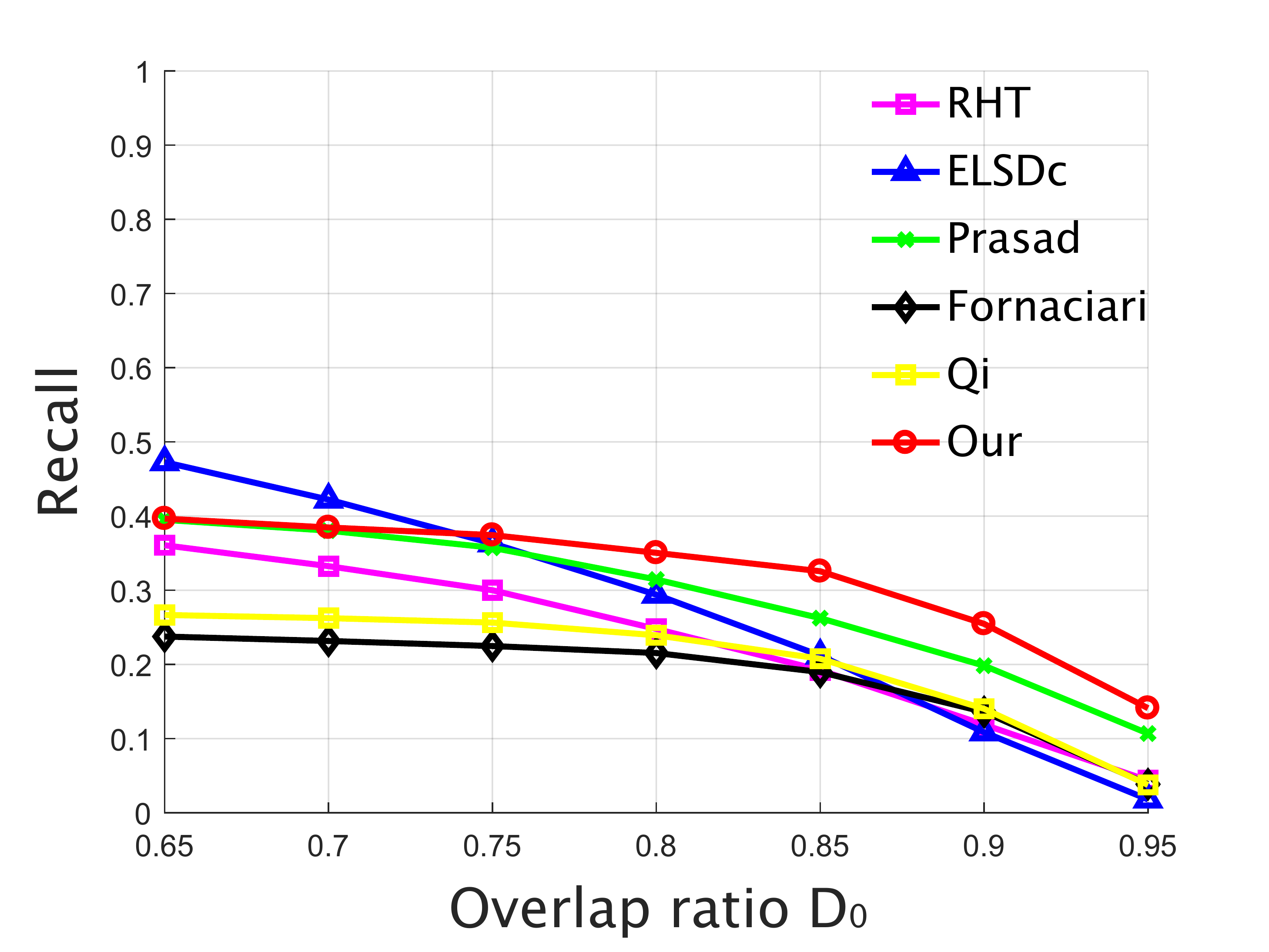}
\label{fig8:Prasad-R}}
\subfigure[]{\includegraphics[width=0.3\hsize]{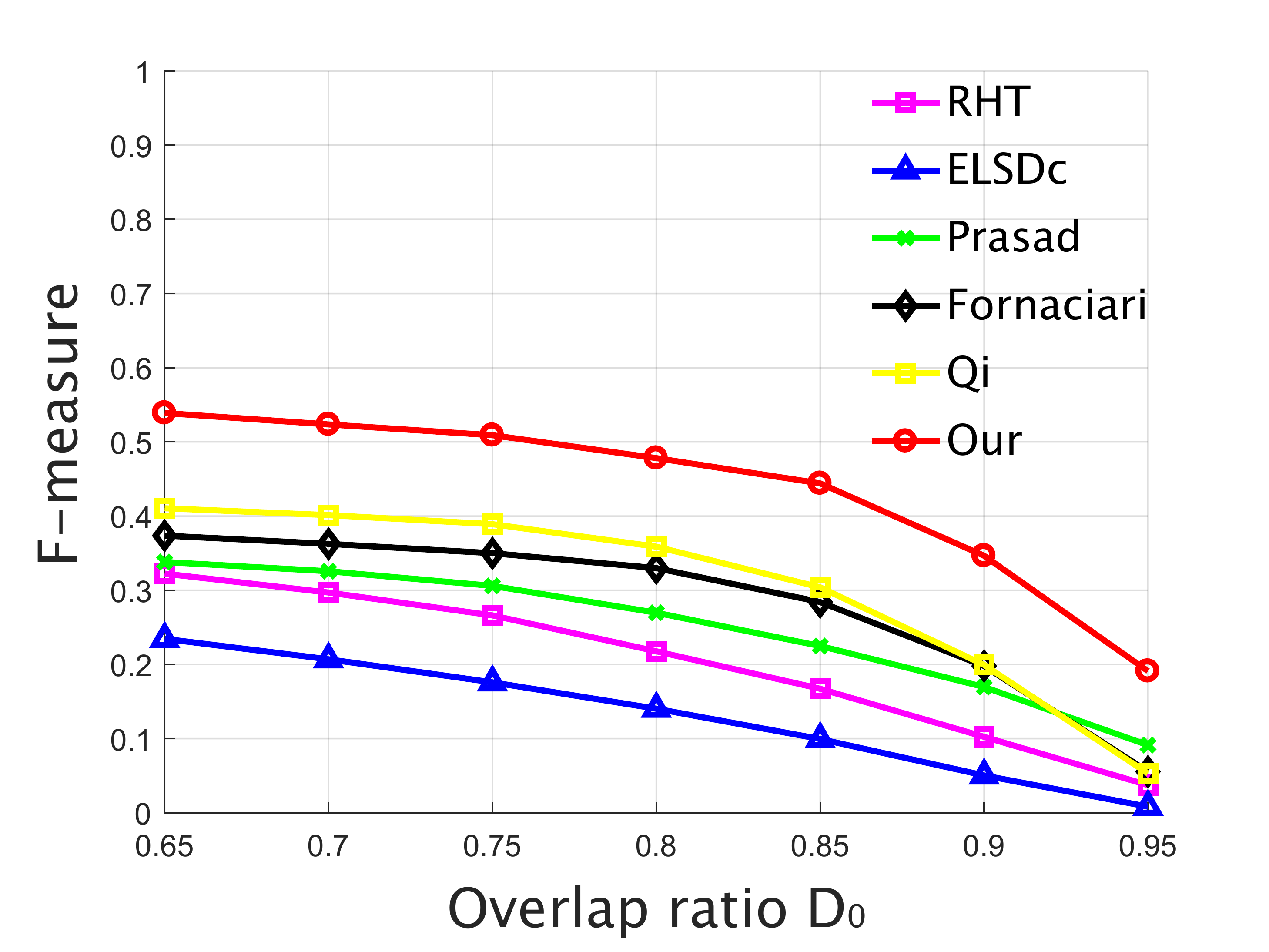}
\label{fig8:Prasad-F}}
\subfigure[]{\includegraphics[width=0.3\hsize]{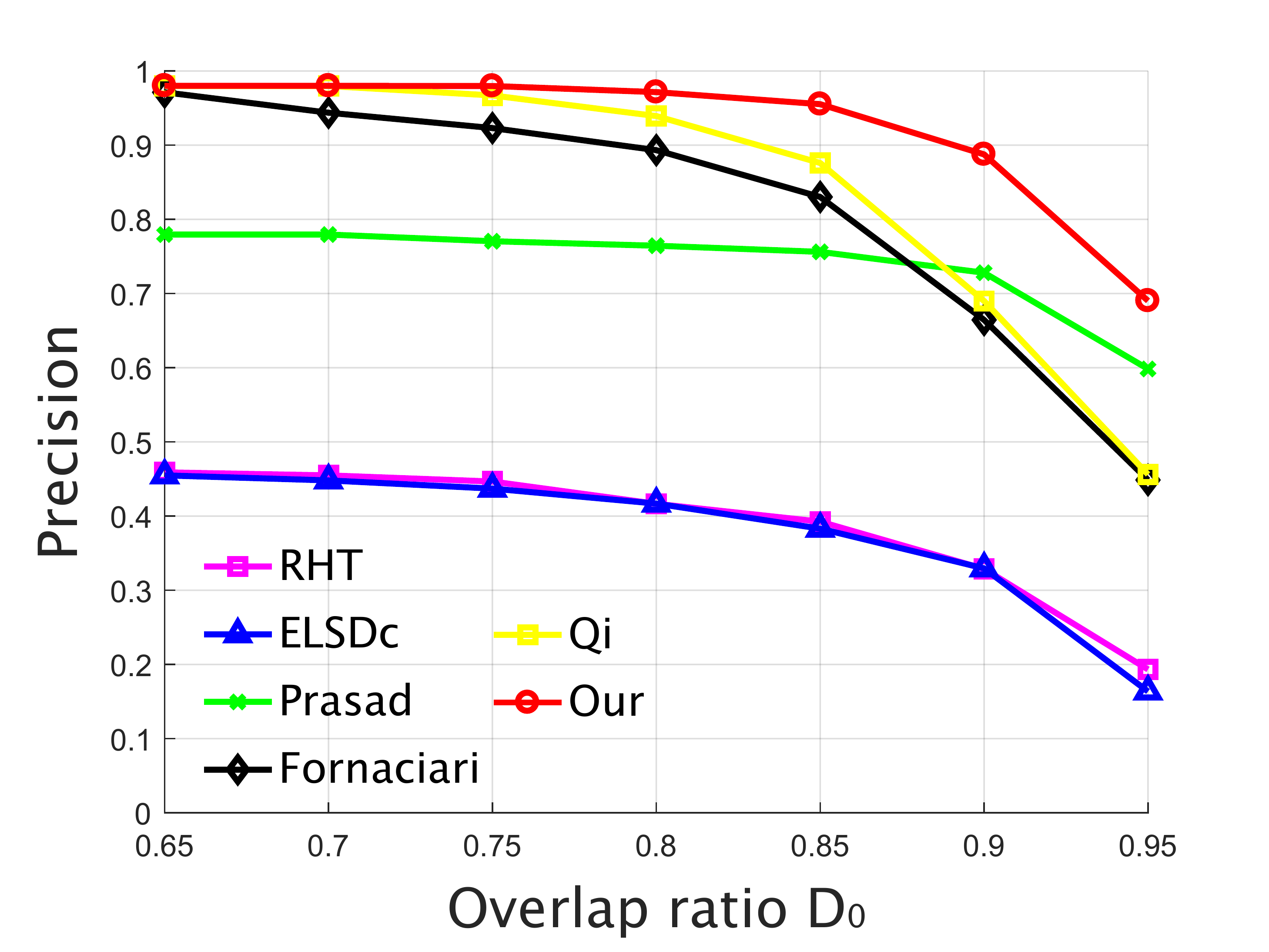}
\label{fig8:Industrial-P}}
\subfigure[]{\includegraphics[width=0.3\hsize]{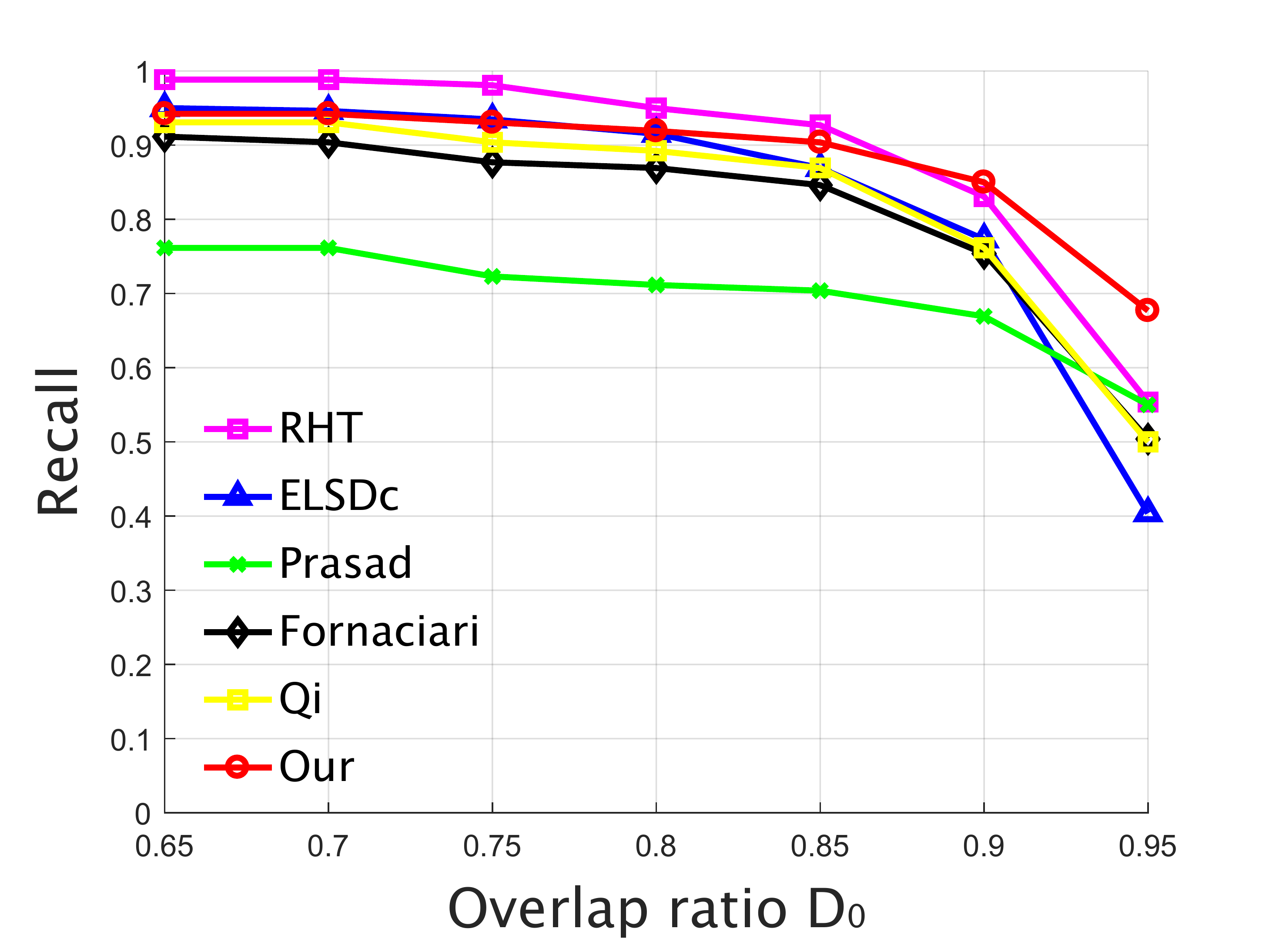}
\label{fig8:Industrial-R}}
\subfigure[]{\includegraphics[width=0.3\hsize]{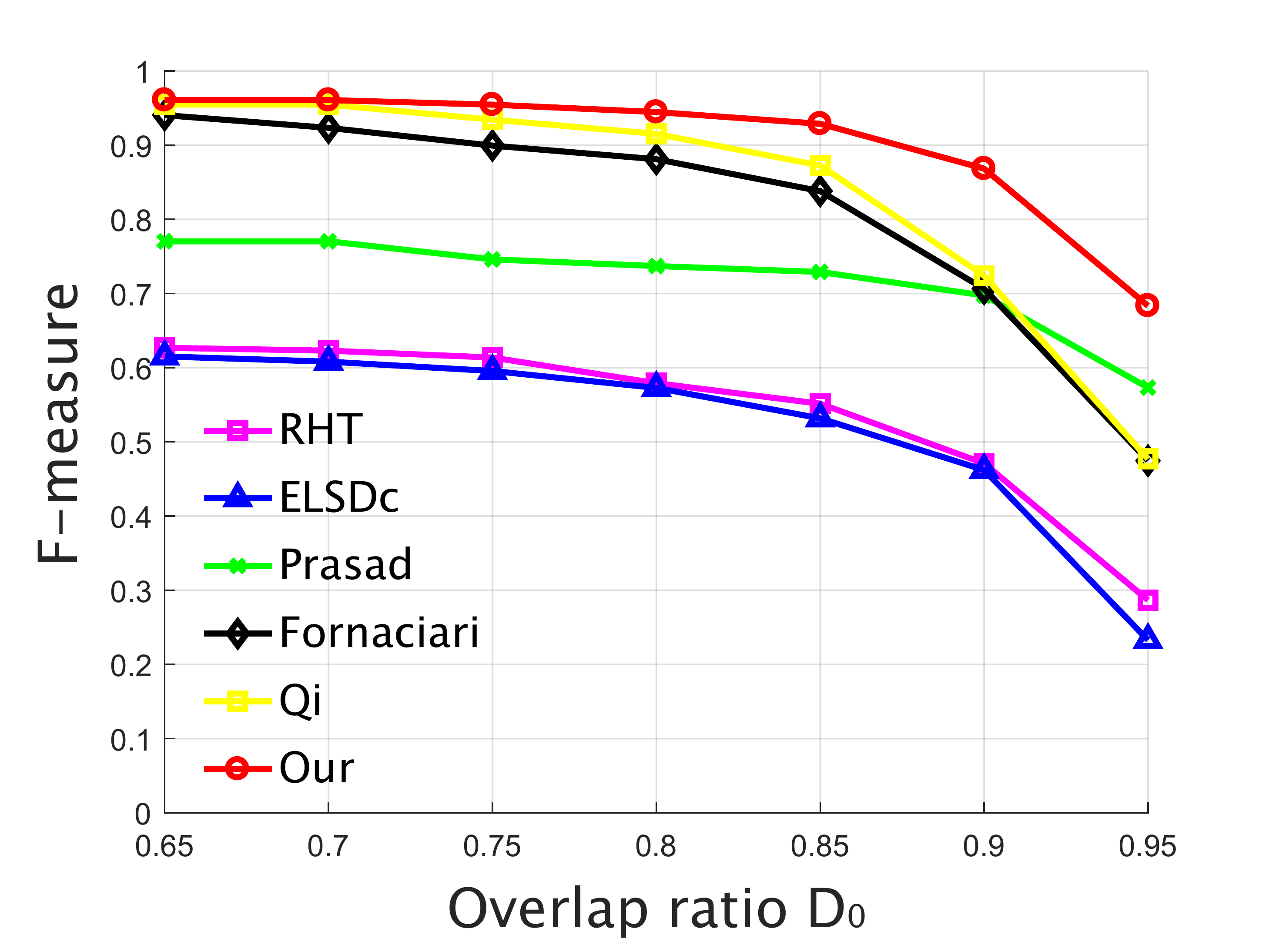}
\label{fig8:Industrial-F}}
\caption{Ellipse detection performance of our method against other methods by varying overlap ratio $D_{0}$ from $0.65$ to $0.95$ at the step of $0.05$ on three real-world datasets. The three rows reveal the metric indexes of precision, recall and F-measure on traffic sign dataset, Prasad dataset and PCB dataset respectively.}
\label{fig8}
\end{figure*}

The detailed ellipse detection results on three real-world datasets are shown in Table \ref{tab1}. The highest scores in precision, recall, F-measure and cost time are stressed in boldface. As we can see, our method achieves the best F-measure scores 0.8959, 0.4781 and 0.9447 across traffic sign dataset, Prasad dataset and PCB image dataset respectively, which highlights its extraordinary overall ellipse detection ability. Moreover, our method holds the winner almost in all precision and recall indexes in three real-world datasets except the recall in PCB dataset which is acquired by RHT \cite{mclaughlin1998randomized}. The higher precision indicates that our ellipse detector is more rigorous to reject false positives and the ellipses reported by our method own larger possibility to truly exist. The precision and recall of our method keep pace, consequently, yielding the larger F-measure scores.
The iterative random search helps RHT \cite{mclaughlin1998randomized} raise the recall, however, its precision and F-measure scores are unsatisfactory due to the lack of novel validations. ELSDc \cite{patraucean2017joint} tries to detect ellipses and line segments simultaneously. However, the side-effects are also brought into such as the limited ellipse detection performance and long computation time in real-world images. Actually, ELSDc \cite{patraucean2017joint} are more suitable to deal with PCB images as its performance gets promoted vastly in PCB dataset. Similar to ELSDc \cite{patraucean2017joint}, Prasad method \cite{prasad2012edge} also suffers from long computation time due to the heavy detection procedures.
From Table \ref{tab1}, it can observe that Qi et al. method \cite{jia2017fast} which is developed on the basis of Fornaciari et al. method \cite{fornaciari2014fast} could achieve second best overall ellipse detection results across three real-world datasets. Moreover, Qi et al. method \cite{jia2017fast} is able to consume quite small running time. The reasons behind this are mainly due to its usage of projective invariant for effectively pruning straight edges, fast arc selection strategy, and simple clustering \cite{jia2017fast, Prasad2010Clustering}. In contrast to Qi et al. method accelerating detection speed at the risk of generating duplicates surrounding a common ground truth, our method employs a more useful and yet relatively more time-consuming hierarchical clustering method for ellipse candidates, which could reduce the false positives significantly. Although our method implemented in MATLAB would further slow down the ellipse detection speed, it could achieve competitive running time compared with the methods proposed by Qi et al. \cite{jia2017fast} and Fornaciari et al. \cite{fornaciari2014fast} which are in C++.\par

Some ellipse detection examples on three real-world datasets are shown in Fig. \ref{fig7}. The images of rows (1) to (3) are from the traffic sign dataset, where exist the disturbances of illumination, varied eccentricity and extremely close concentric ellipses. The second three-row images are from Prasad dataset which is the most complicated among three real-world datasets due to noise, occlusions and various backgrounds. The left three images are from PCB dataset, which is with the substantial Gaussian white noise and blur. As a result, both accurately and efficiently detecting ellipses in such images is difficult to an ellipse detector. As illustrated in Fig. \ref{fig7}, RHT \cite{mclaughlin1998randomized} tends to detect every possible ellipse while generates many false positives. The detection performance of RHT gets worse especially in the images with substantial noise and textures. Although ellipse detectors proposed by Fornaciari et al. \cite{fornaciari2014fast} and Qi et al. \cite{jia2017fast} are very efficient, both methods report many ellipse duplicates as well as resulting in poor location accuracy, as revealed in the images of (2) to (5) rows. In contrary, Prasad method \cite{prasad2012edge} and ELSDc \cite{patraucean2017joint} could relatively more accurately locate the ellipses. The relating concerns of both Prasad method and ELSDc are the massive missing detections for ground truth and modeling small contours as ellipses, which worsens the overall ellipse detection performance as the F-measure scores shown in Table \ref{tab1}. Unlike other ellipse detectors being tough to balance the issues between accuracy and efficiency, our method can both efficiently and accurately detect the ellipses and is robust to noise and textures. Especially, our method also performs well in handling the incomplete and occluded ellipses, as shown in the last column of Fig. \ref{fig7}. The elements guarantee our method's accuracy mainly due to the false control ability of arc-support LSs, novel verification criteria and self-calibrated refinement. And the reasons for the good efficiency are attributed to arc-support LSs alleviating the disturbance of straight LSs and effective initial ellipse generation aided with polarity constraint, region restriction and adaptive inliers criterion. \par

In order to comprehensively evaluate the ellipse detection performance of the compared methods, we vary the overlap ratio threshold $D_{0}$ from $0.65$ to $0.95$ at the step of $0.05$, the higher of which indicates the stricter a detected ellipse being regarded as true positive. The corresponding results are shown in Fig. \ref{fig8}. Again, our method achieves the best overall ellipse detection performance among three real-world datasets as the F-measure curves are above those of compared methods, which accords with the before performance analysis. It is evident that our method shows high-quality ellipse detection performance.

\subsection{Localization Accuracy and Efficiency Analysis}

\begin{table}[!tb]
  \centering
  \caption{The Mean Overlap Ratio (MOR) of Correctly Detected Ellipses on Three Real-world Datasets.}
  \label{tab2}
  \begin{tabular}{cccc}
  \toprule[1pt]
  MOR & Traffic Sign & Parasad Dataset & PCB Dataset \\\midrule[1pt]
  RHT \cite{mclaughlin1998randomized}    & 0.9080 & 0.8963 & 0.9459 \\
  ELSDc \cite{patraucean2017joint}       & 0.9229 & 0.8818 & 0.9352 \\
  Prasad \cite{prasad2012edge}           & 0.9226 & 0.9144 & \textbf{0.9603} \\
  Fornaciari \cite{fornaciari2014fast}   & 0.9274 & 0.9080 & 0.9442 \\
  Qi \cite{jia2017fast}                  & 0.9239 & 0.9047 & 0.9428 \\
  Our                                    & \textbf{0.9383} & \textbf{0.9291} & 0.9574 \\\bottomrule[1pt]
  \end{tabular}
\end{table}

\begin{figure}[!tb]
\centering
\subfigure[]{\includegraphics[width=0.45\hsize]{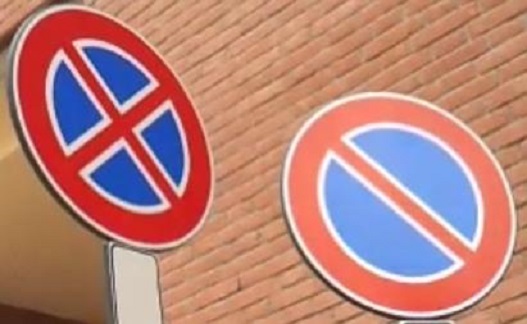}
\label{fig9:accuracy1}}
\subfigure[]{\includegraphics[width=0.45\hsize]{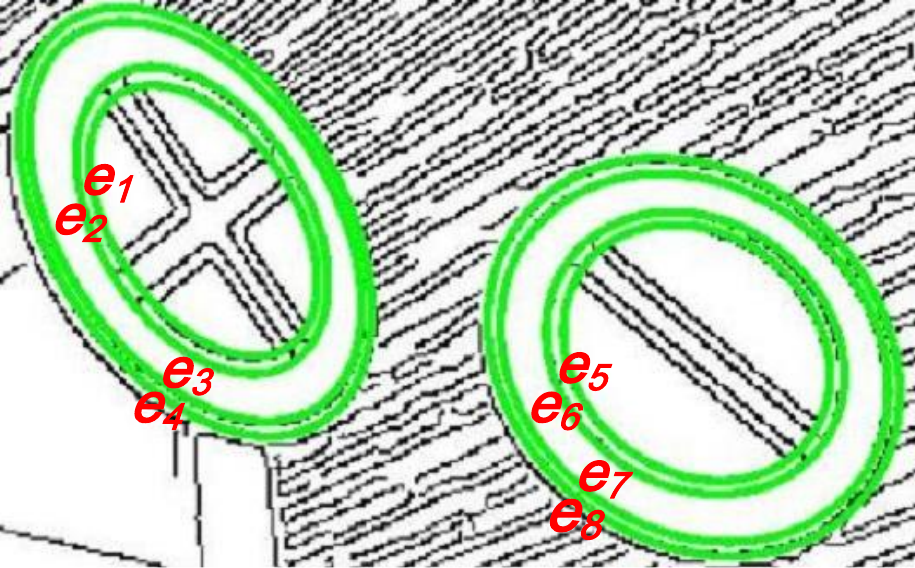}
\label{fig9:accuracy2}}
\subfigure[]{\includegraphics[width=0.45\hsize]{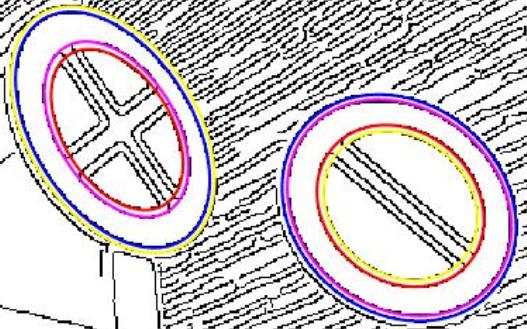}
\label{fig9:accuracy3}}
\subfigure[]{\includegraphics[width=0.45\hsize]{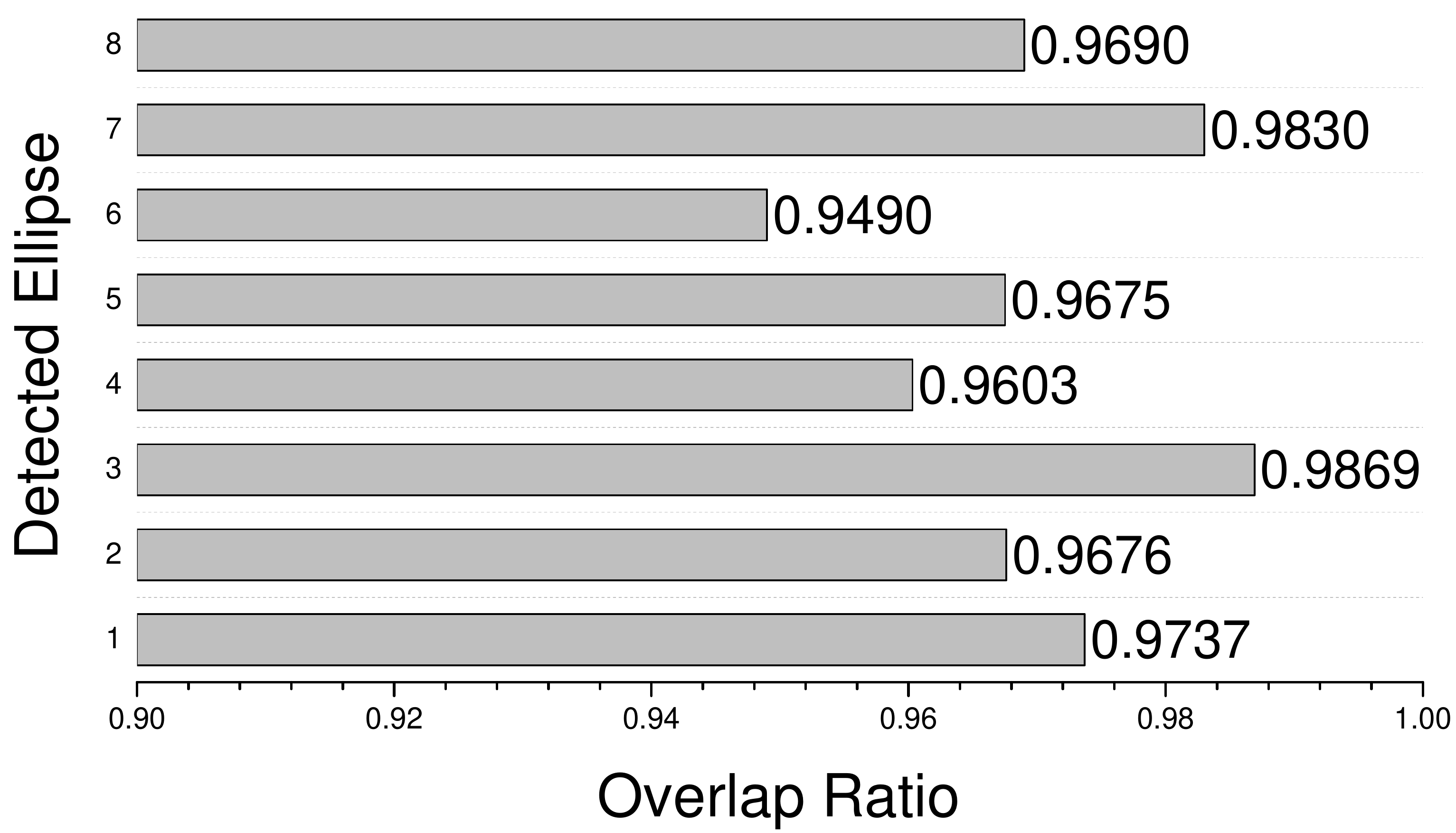}
\label{fig9:accuracy4}}
\caption{An illustration example of the proposed method with high localization accuracy. (a) input image; (b) ground truth; (c) eight detected ellipses; (d) the overlap ratio of each detected ellipse with ground truth.}
\label{fig9}
\end{figure}

Localization accuracy is a critical index to testify whether an ellipse detector to be high-quality or not. To this end, we compute each ellipse detection method's mean overlap ratio (MOR) of correctly detected ellipses on three real-world datasets and the results are shown in Table \ref{tab2}. Our method achieves the best MOR scores in traffic sign dataset and Prasad dataset and second highest MOR $0.9574$ in PCB dataset compared to $0.9603$ which is acquired by Prasad method \cite{prasad2012edge}. The higher MOR indicates that our method aims at the high localization accuracy and does not rest content with picking out the true positives, which stands the proposed method out the compared methods. Actually, accurate localization could favor an ellipse detector to distinguish the very closed ellipses. An accurate ellipse detection example of our method is shown in Fig. \ref{fig9}. There are eight ground truth in the input image and the average distance of each two concentric ellipses is $\Delta 4.18_{-1.08}^{+0.78}$ pixels. Although the ellipses are so close, our method still can successfully locate each individual and report high overlap ratio, as shown in Fig. \ref{fig9:accuracy3} and Fig. \ref{fig9:accuracy4}.\par

\begin{figure}[!tb]
\centering
\includegraphics[width=0.9\hsize]{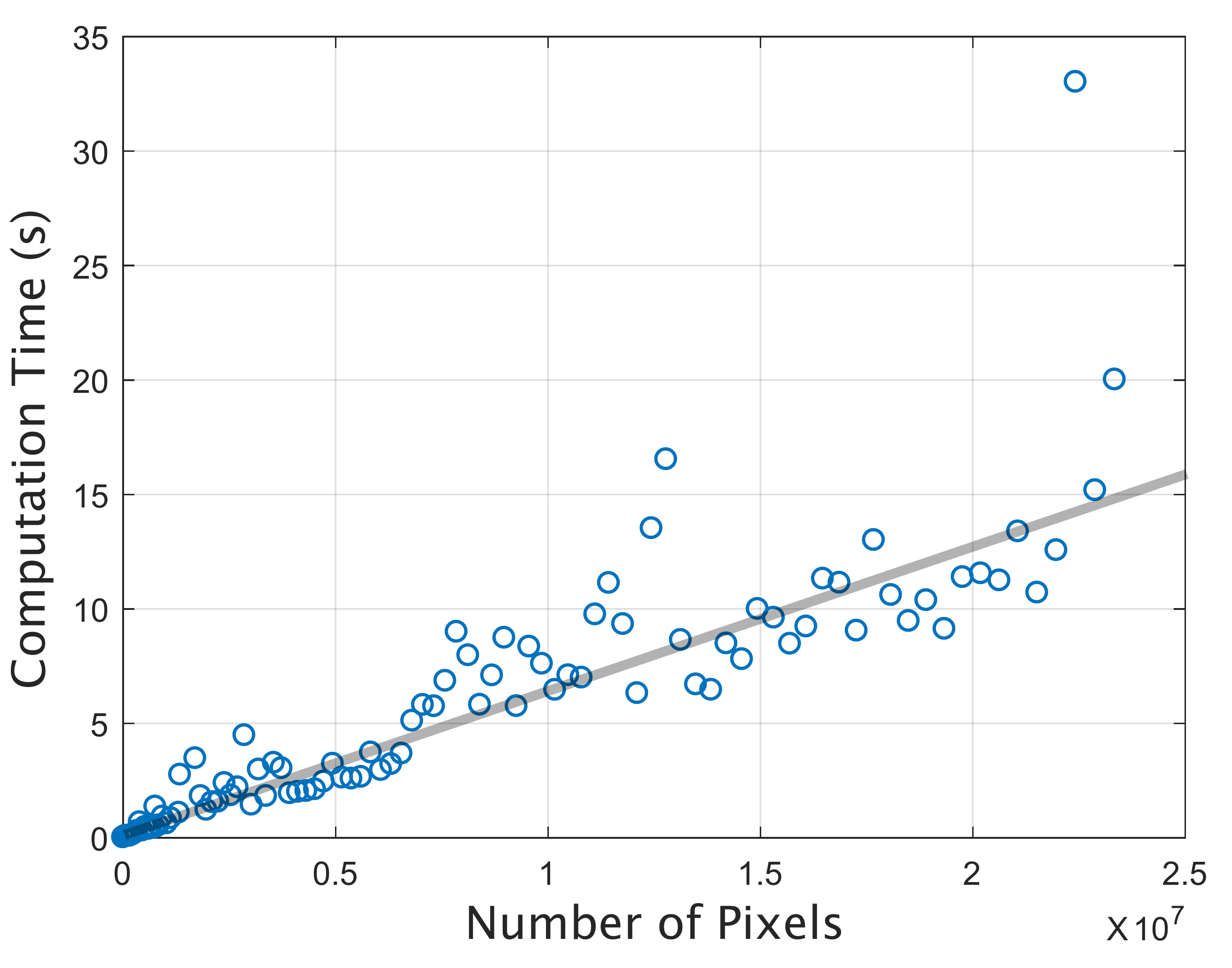}
\caption{Computation time with regard to the number of pixels of real-world images.}
\label{fig10}
\end{figure}

In order to verify the quadratic complexity of the proposed ellipse detector in image longer length $N$ and arc-support groups $N_{G}$, we record the computation time of 100 different real-world images, the sizes of which range from 46 x 51 to 4600 x 5100. The correlation between computation time and the number of pixels is shown in Fig. \ref{fig10}. The scatters are general in linear distribution and the ellipse computation time almost linearly increases with the number of pixels, which indicates that our method is quadratic in $N$ ($N>N_{G}$ in most of the images). Admittedly, the ellipse detector should own larger time complexity than line segment or circle detector. Our method is still efficient and can handle the real-world images in quadratic time complexity which is superior to most of existing ellipse detectors.

\subsection{Robustness to Parameters Setting and Ellipse Variations}
\begin{figure}[!t]
\centering
\subfigure[]{\includegraphics[width=0.48\hsize]{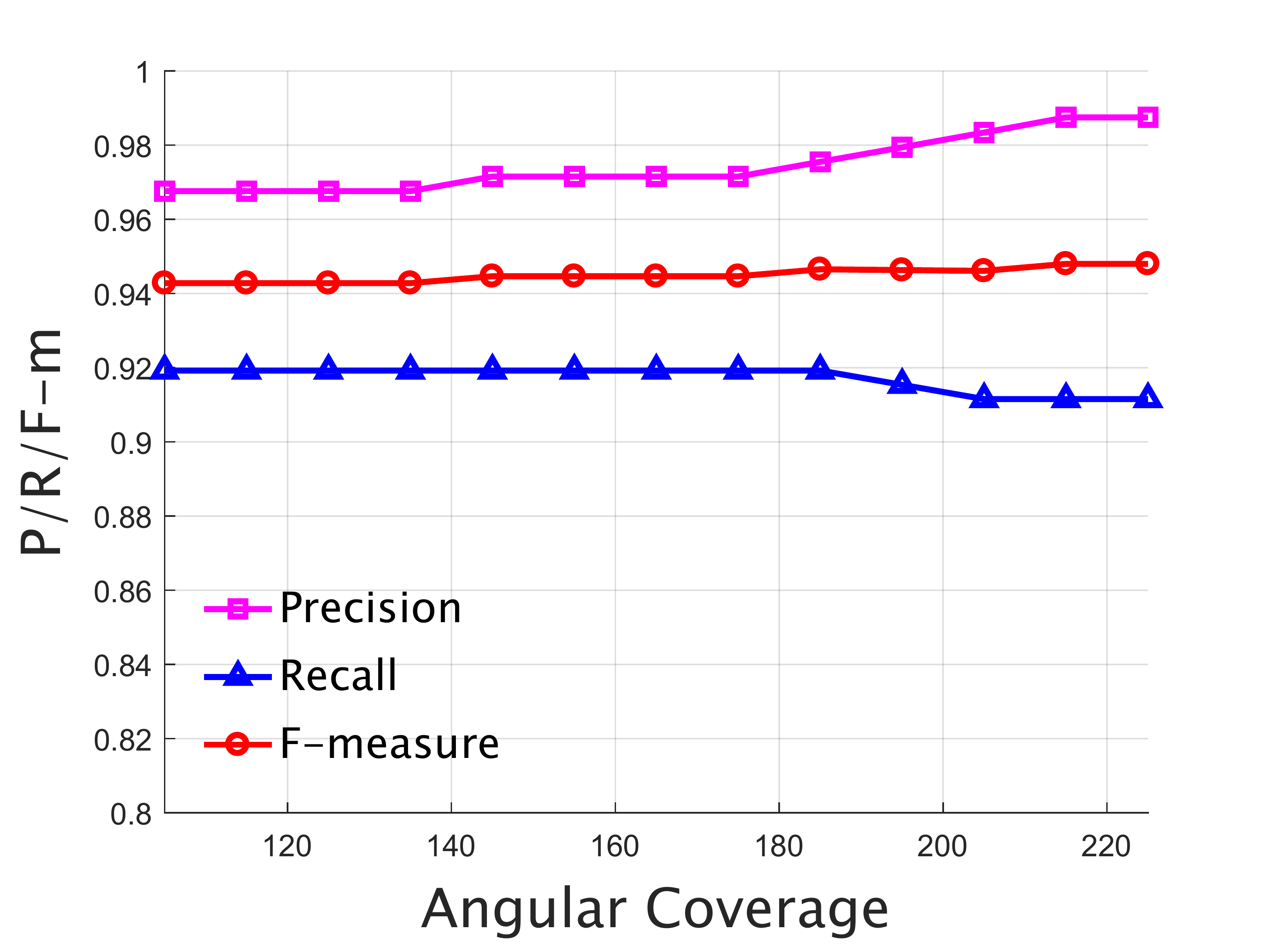}
\label{fig11:AngleVarying}}
\subfigure[]{\includegraphics[width=0.48\hsize]{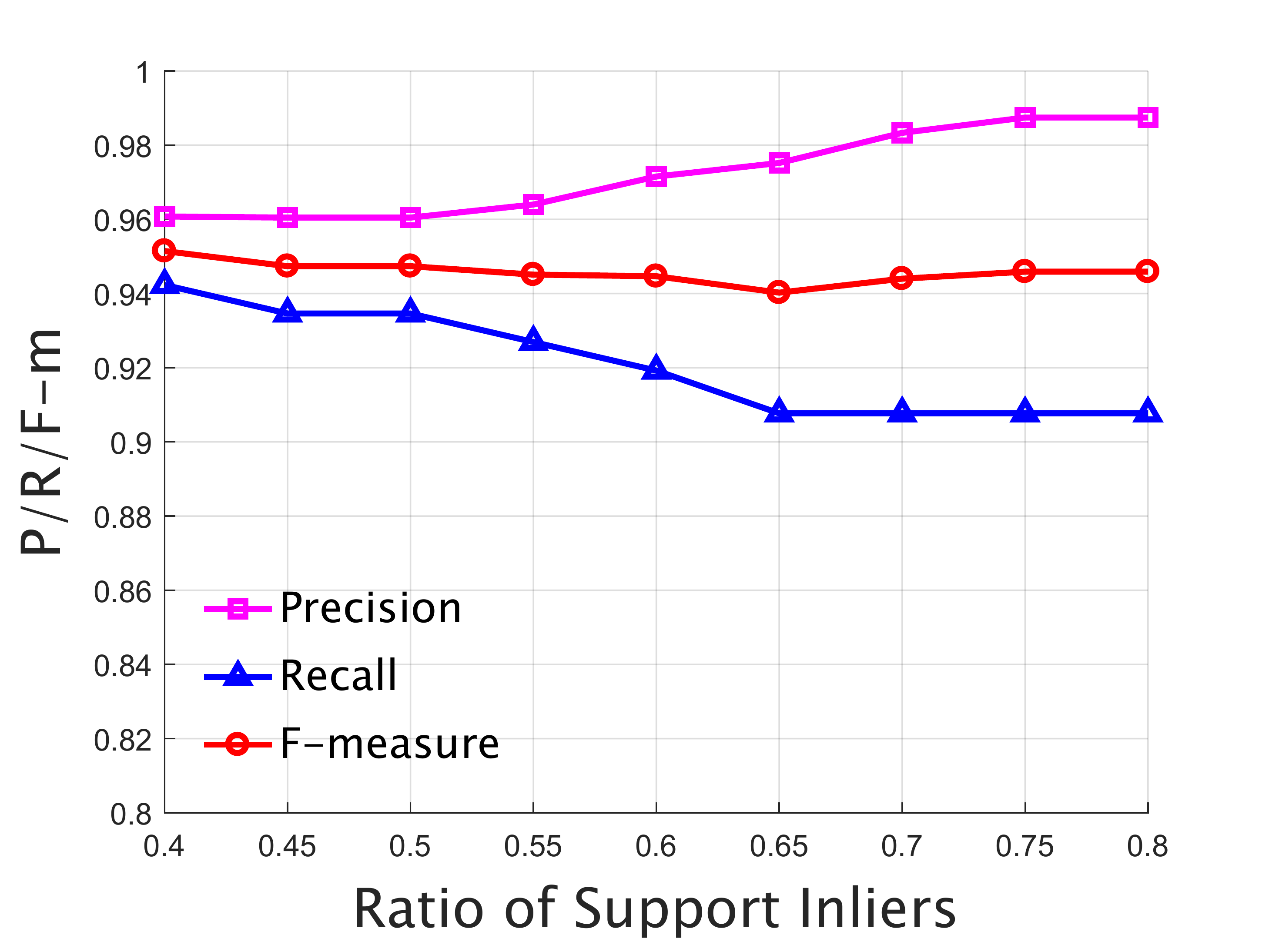}
\label{fig11:TrVarying}}
\caption{Ellipse detection performance of the proposed method in PCB image dataset with varying angular coverage and ratio of support inliers. a) the elliptic angular coverage are set from $105^\circ$ $\sim$ $225^\circ$ at step of $10^\circ$ with fixed ratio of support inliers 0.6; b) the ratio of support inliers ranges from 0.4 $\sim$ 0.8 at step of 0.05 while the angular coverage is $165^\circ$.}
\label{fig11}
\end{figure}

The angular coverage $T_{ac}$ and the ratio of support inliers $T_r$ are two extrinsic parameters of the proposed ellipse detection method. Firstly, $T_{ac}$ and $T_{r}$ have geometric significance, which enables us easy to tune when applied in the real application. Secondly, both parameters are insensitive and have robustness in a wide setting range. To validate the robustness to tunable parameters, we select PCB dataset as the testset and perform quantitative experiments. We first freeze $T_r$ as $0.6$ which is the default parameter and vary the elliptic angular coverage $T_{ac}$ from $105^\circ$ to $225^\circ$ at the step of $10^\circ$. Then the curves of precision, recall and F-measure according to the experimental results are plotted, as shown in Fig. \ref{fig11:AngleVarying}. Similarly, the ratio of support inliers $T_r$ is changed from $0.4$ to $0.8$ at the step of $0.05$ and the angular coverage $T_{ac}$ are fixed to the default parameter $165^\circ$. The corresponding ellipse detection performance is shown in Fig. \ref{fig11:TrVarying}. As the angular coverage $T_{ac}$ and ratio of support inliers $T_r$ rise, the recall tends to decline as the detected ellipses are more likely to be rejected due to the stricter requirements. However, the precision gets boosted since the detected ellipses are purer. Notably, the F-measure curves in both Fig. \ref{fig11:AngleVarying} and Fig. \ref{fig11:TrVarying} are relatively smooth and little fluctuating in a wide range, which reveals the robustness of the proposed method to different parameter settings.\par

\begin{figure}[!t]
\centering
\subfigure[]{\includegraphics[width=0.48\hsize]{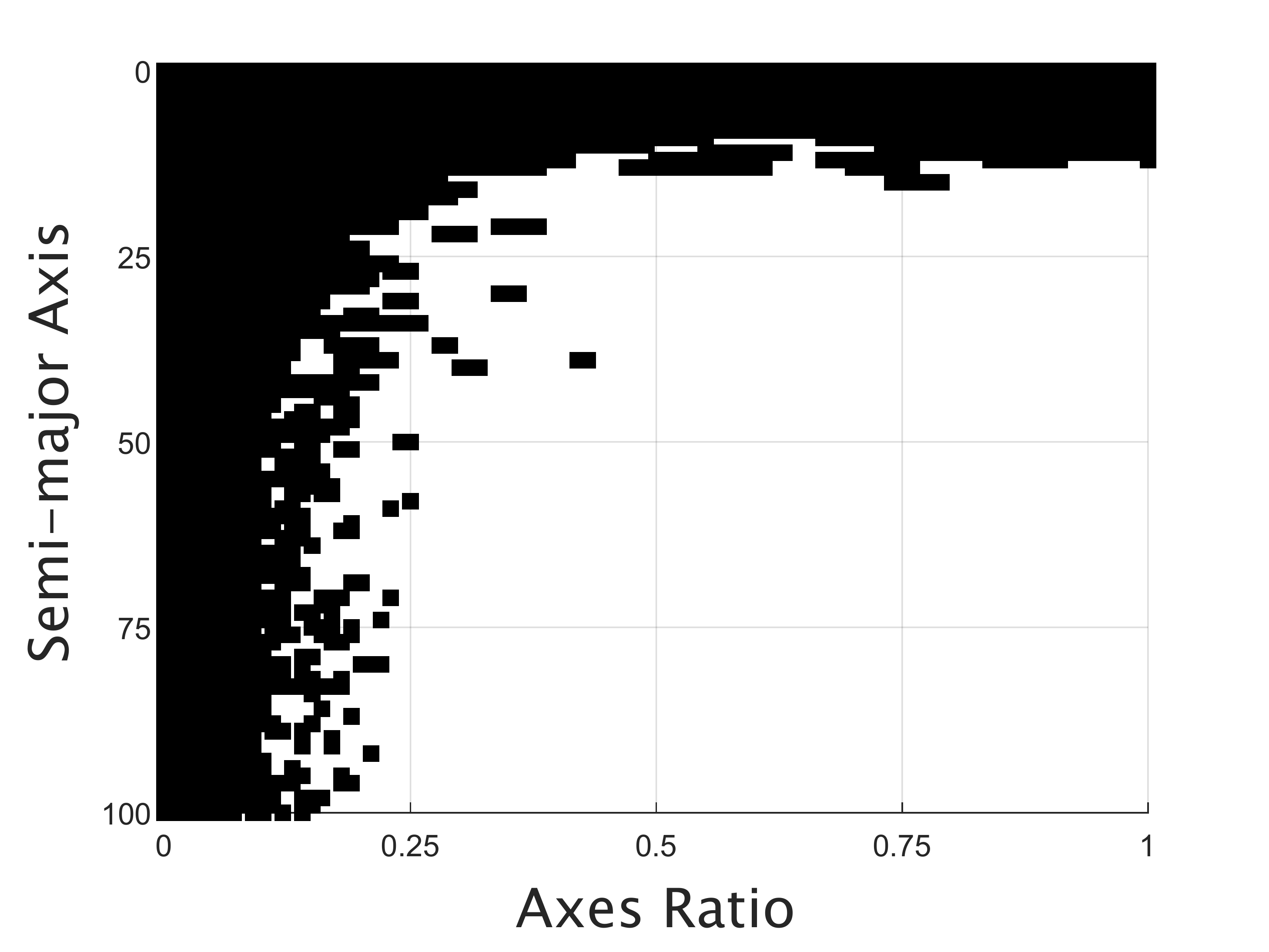}
\label{fig12:a}}
\subfigure[]{\includegraphics[width=0.48\hsize]{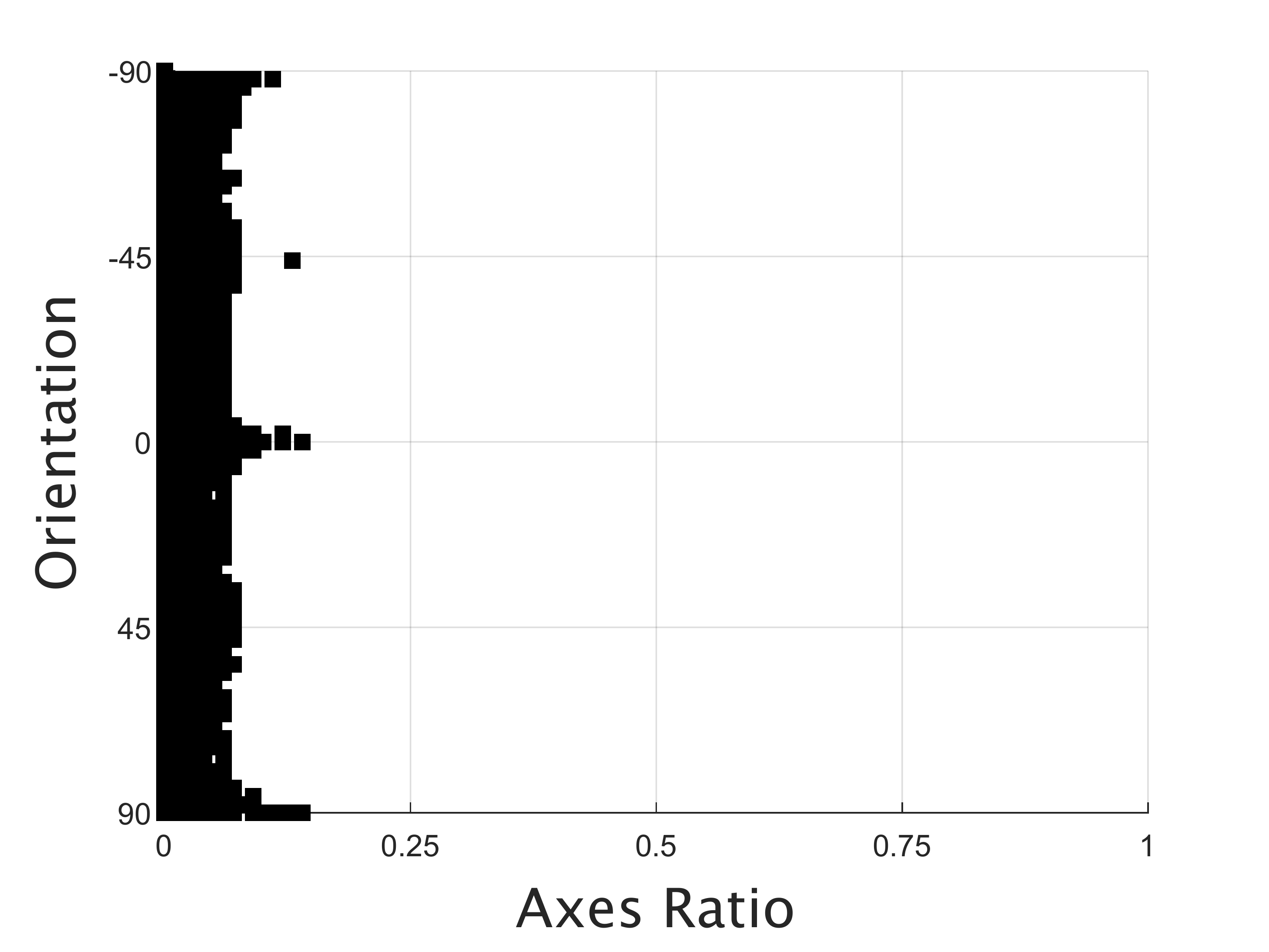}
\label{fig12:b}}
\subfigure[]{\includegraphics[width=0.48\hsize]{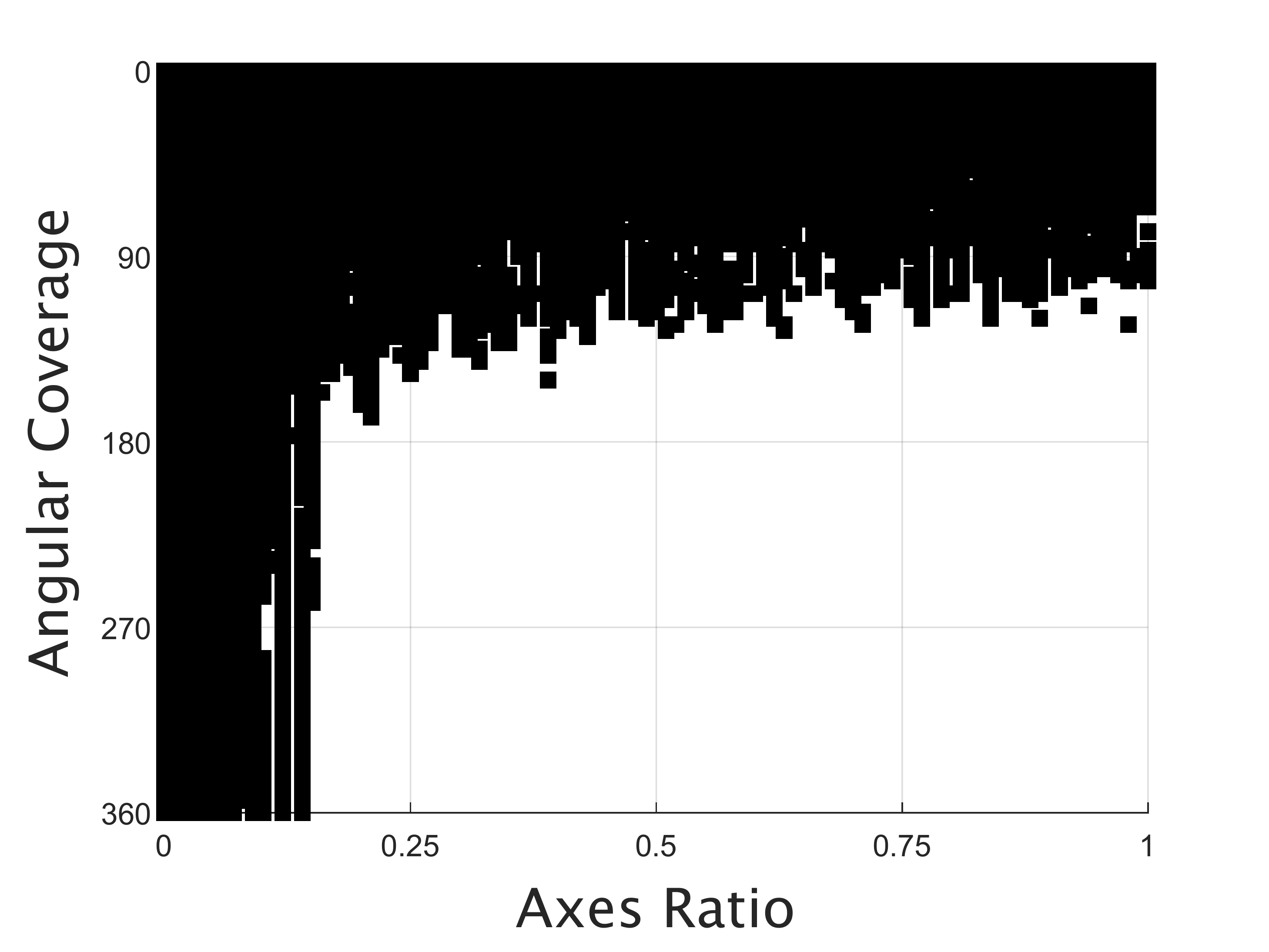}
\label{fig12:c}}
\subfigure[]{\includegraphics[width=0.48\hsize]{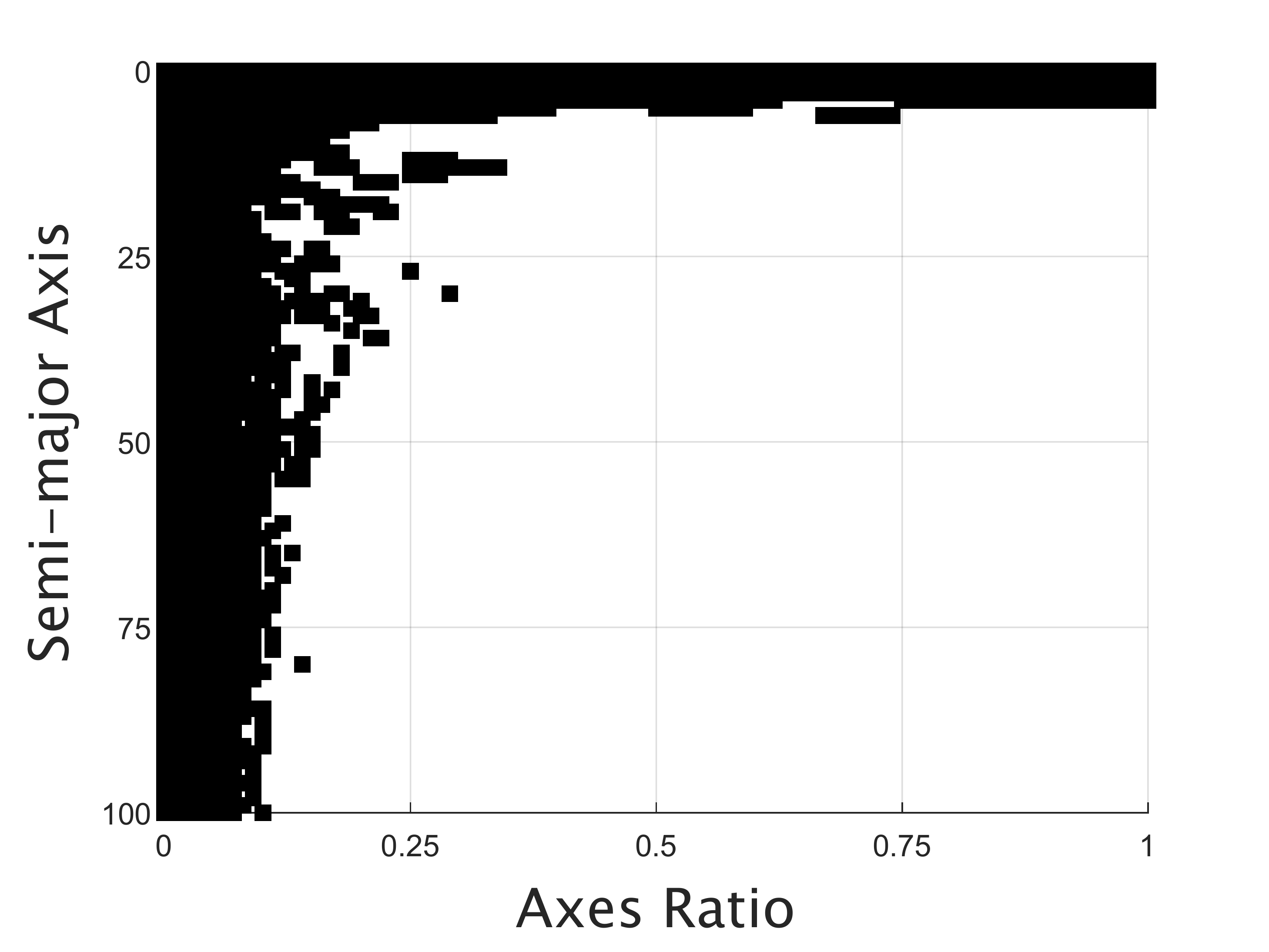}
\label{fig12:d}}
\caption{Extensive detection results subject to various ellipse variations. The horizontal axis is the ratio of semi-minor axis to semi-major one, which ranges from 0.01 to 1 at the step 0.01. The vertical axes of (a), (b) and (c) are the semi-major axis length, ellipse orientation and angular coverage of ellipse arc. (d) shows the effects after upscaling the synthetic images two times.}
\label{fig12}
\end{figure}

In order to investigate the robustness of our method to different ellipse variations such as ellipse size, orientation and incompleteness, three synthetic datasets are prepared. The first dataset includes 10000 images, in which the semi-major axis of the ellipse is varied from $1\sim 100$ pixels at the step of 1 pixel and the axes ratio ranges from 0.01 to 1 at the step of 0.01. To evaluate the influence of orientation, we build the second dataset by rotating the ellipse from $-88^\circ$ to $90^\circ$ at the step of $2^{\circ}$. For each orientation, the major-axis is fixed to 100 pixels and the axes ratio changes from $0.01\sim 1$ at the step of 0.01, which totally results in 9000 images. Actually, a high-quality ellipse detector should accurately detect the incomplete ellipses, namely elliptic arcs. Therefore, the third dataset is built and consists of 12000 images, where the angular coverage of ellipse varies from $3^\circ$ to $360^\circ$ at the step of $3^\circ$ and the axes ratio ranges from $0.01 \sim 1$ at the step of $0.01$. Each synthetic image contains an ellipse and is with the size of 250 x 250.\par

The effects of ellipse variations on our ellipse detector are shown in Fig. \ref{fig12}, where the white region indicates the corresponding ellipses could be correctly detected while the black region means the detection failures. Firstly, in Fig. \ref{fig12:a}, our method has wide successful detection area and could detect the small ellipse with the semi-major axis of about 20 pixels and axes ratio of 0.25. The extremely oblate and small ellipses are failed to detect by our method as well as by the methods proposed by Fornaciari et al. \cite{fornaciari2014fast} and Qi et al. \cite{jia2017fast}. Secondly, the black region distributed vertically in Fig. \ref{fig12:b}, which indicates that the ellipse detection performance is invariant to ellipse orientation. Our method is robust to the orientation as it is a basic nature of high-quality ellipse detector. Thirdly, our method can successfully detect the elliptic arc with angular coverage of about $165^\circ$ since our parameter $T_{ac}$ is acquiescently set to $165^\circ$, as shown in Fig. \ref{fig12:c}. This result reveals that our method is able to tackle the incomplete ellipses and detect the specified elliptic arc with assigned angular coverage. Finally, our ellipse detection performance gets improved after upscaling the image size two times since the black region shrinks in Fig. \ref{fig12:d}, which provides a feasible approach to boost detection performance of the proposed ellipse detector.

\subsection{Polarity-specific Ellipse Detection}
\begin{figure}[!t]
\centering
\subfigure[]{\includegraphics[width=0.48\hsize]{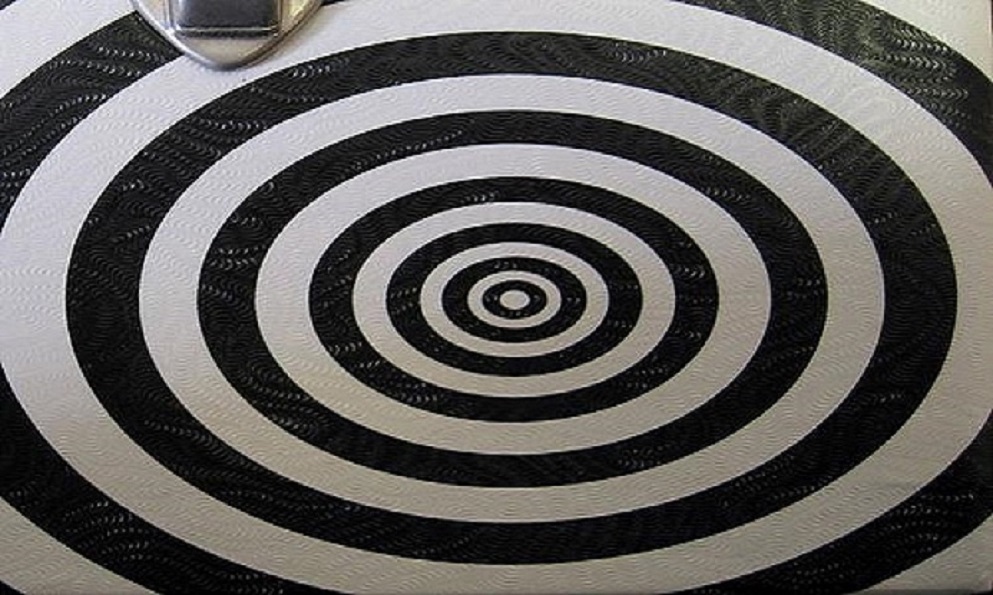}
\label{fig13:a}}
\subfigure[]{\includegraphics[width=0.48\hsize]{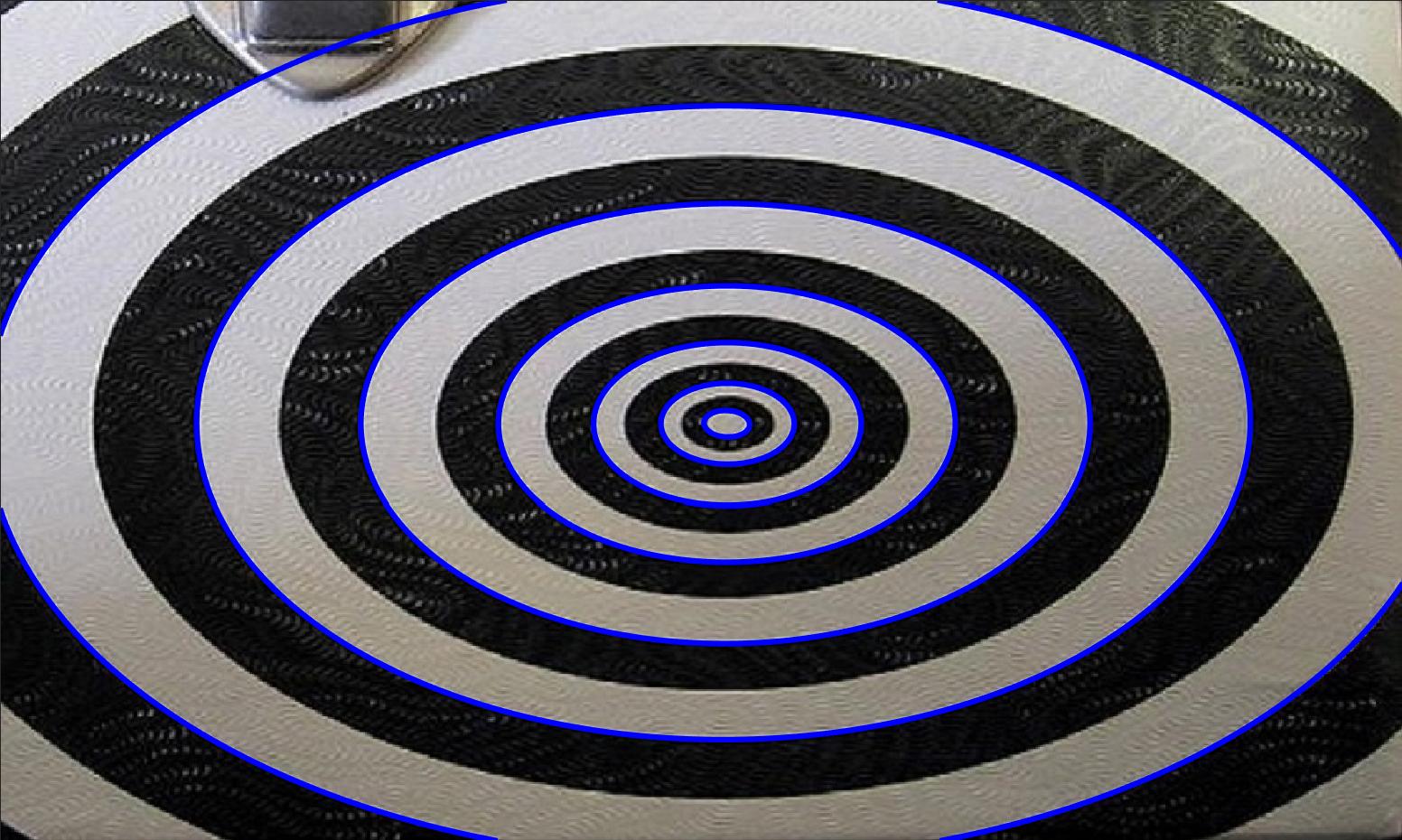}
\label{fig13:b}}
\subfigure[]{\includegraphics[width=0.48\hsize]{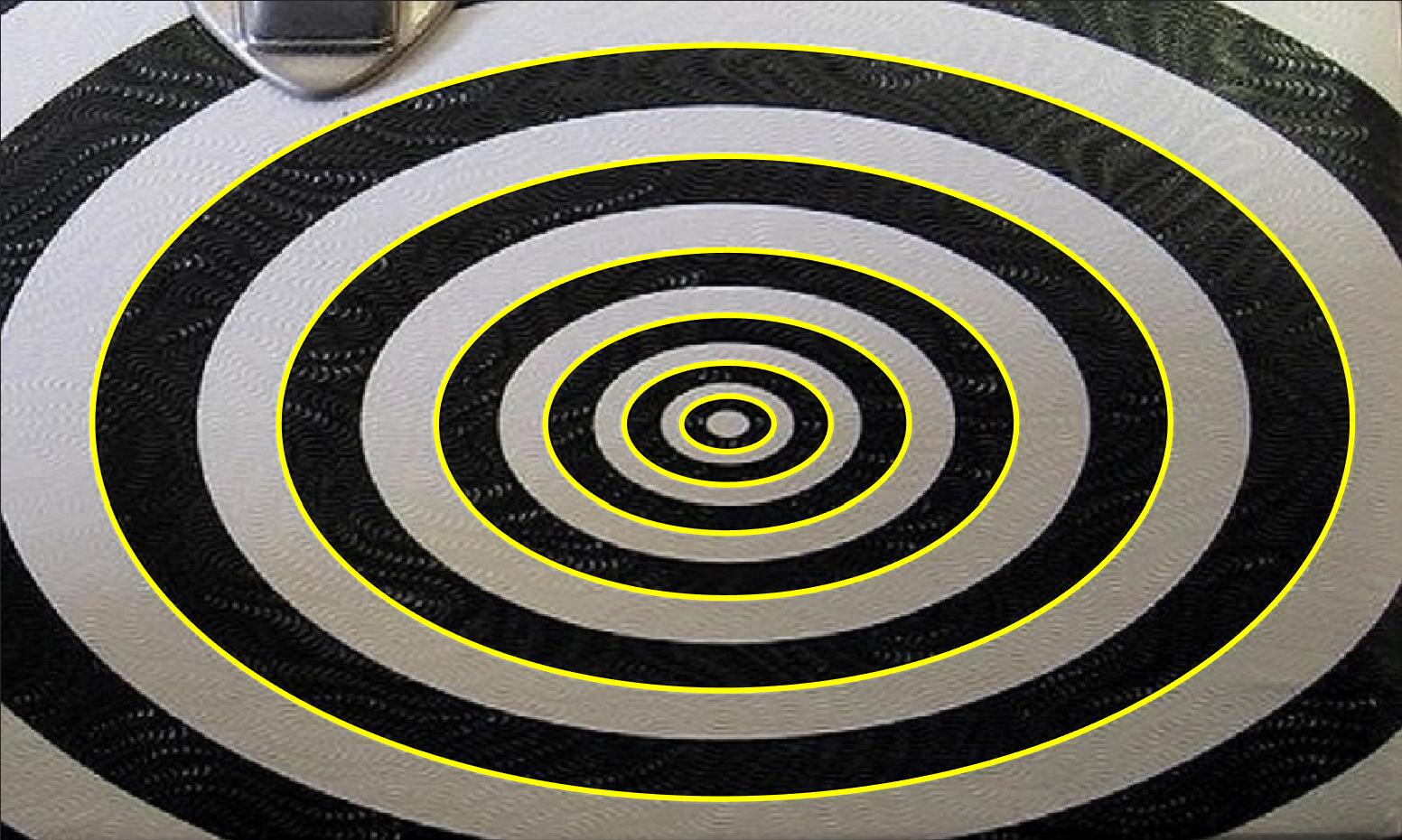}
\label{fig13:c}}
\subfigure[]{\includegraphics[width=0.48\hsize]{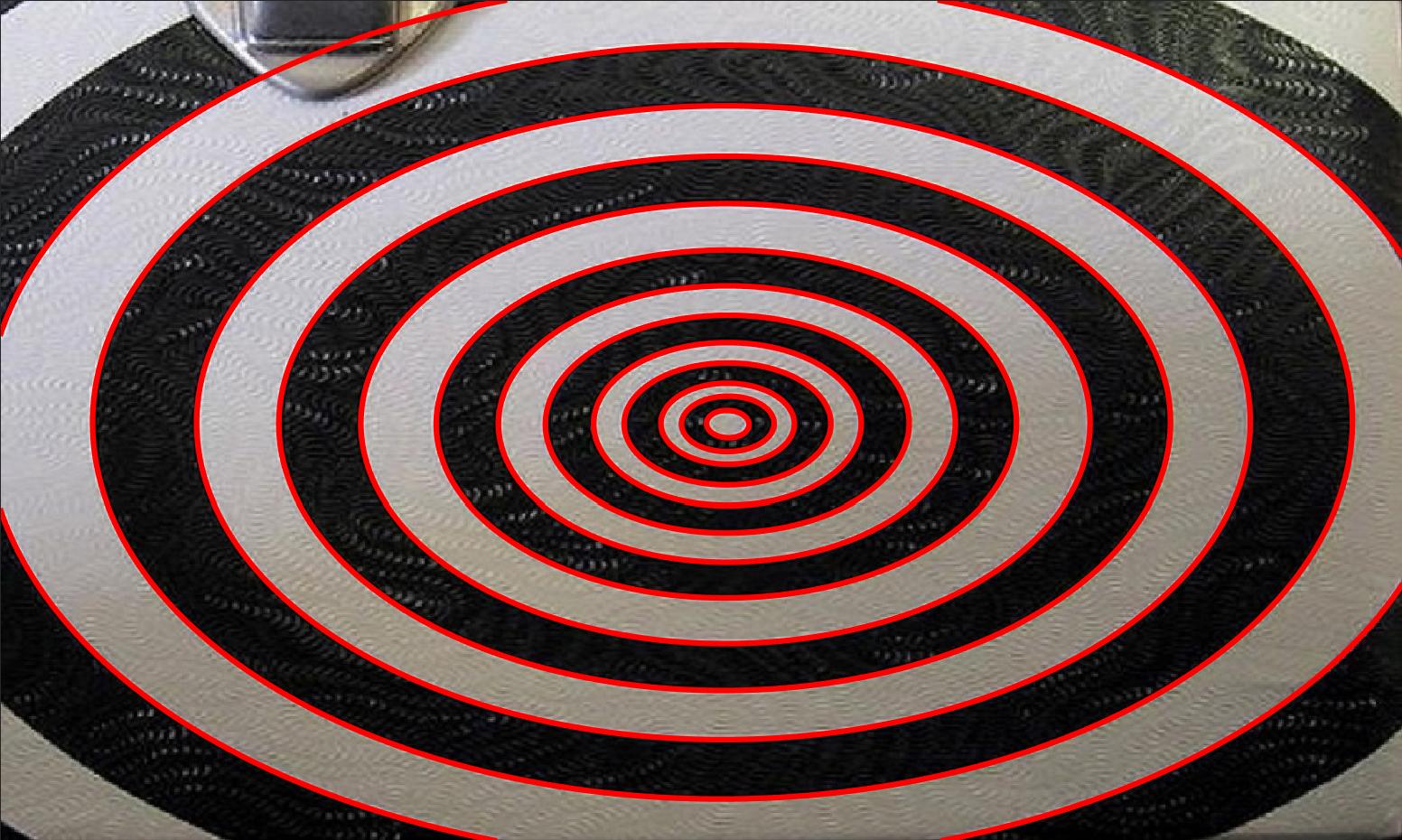}
\label{fig13:d}}
\caption{An example of polarity-specific ellipse detection. (a) origin image, 993 x 595; (b) detection for the ellipse whose polarity is positive; (c) detection for the negative polarity ellipse; (d) detecting all ellipses in the image.}
\label{fig13}
\vspace{-3pt}
\end{figure}

Recall that the polarity of an ellipse is positive if the corresponding inside adjacent area of the boundary is brighter than outside, otherwise is negative. Actually, our method is able to detect the polarity-specific ellipses because we only need to retain the arc-support LSs with the corresponding polarity for generating the initial ellipse set. As shown in Fig.~\ref{fig13:a}, the black elliptic ring belts and white ring belts are concentric and adjacent. Each ring belt will generate two different ellipses with positive or negative polarity. In Fig.~\ref{fig13:b}, the concentric ellipses with positive polarity are successfully detected by our method and they are highlighted in blue color. In Fig.~\ref{fig13}(c), the detected ellipses in yellow are all with negative polarity. Naturally, if we use all the arc-support LSs for ellipse detection, the target is to detect all potential ellipses in the image, as the detected red ellipses in Fig.~\ref{fig13}(d). The information of polarity of arc-support LS is greatly important and useful, which not only contributes to reducing the computation time for searching all the valid paired arc-support groups but also helps to detect the polarity-specific ellipses in the certain case.

\section{Conclusion}\label{sec:conclusion}
In this paper, we propose a high-quality ellipse detection method by introducing the arc-support LSs, which aims at both accurately and efficiently detecting ellipses in real-world images. To this end, our method follows a four-stage ellipse detection framework: arc-support groups forming, initial ellipse set generation, clustering, and candidate verification. With the help of arc-support LSs, straight LSs are filtered and the abundant geometric features such as overall gradient direction of the local area, arc-support direction and polarity can be thoroughly exploited. The robust forming of arc-support groups, the adoption of the superposition principle of ellipse fitting and the efficient generation of initial ellipse set with three novel geometric constraints guarantee the overall efficiency of the proposed method. Moreover, the rigorous ellipse verification defend the high localization accuracy and robustness as well as rejecting the false positives. The self-calibrated refinement facilitates higher accuracy. The quantitative experiments compared with existing novel methods evidently demonstrate that our method could well balance the relationship between accuracy and efficiency, and achieves the high-quality ellipse detection performance.


%





\ifCLASSOPTIONcaptionsoff
  \newpage
\fi



\bibliographystyle{IEEEtran}
\bibliography{refs}
%




%





\begin{IEEEbiography}[{\includegraphics[width=1in,height=1.25in,clip,keepaspectratio]{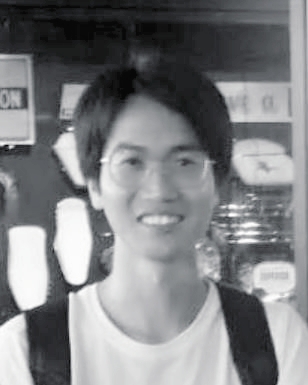}}]{Changsheng Lu}
received the B.S. degree in Automation from Southeast University, Nanjing, China, in 2017. Currently, he is an academic M.S. student with the Key Laboratory of System Control and Information Processing, Shanghai Jiao Tong University. He has wide research interests mainly including computer vision, machine learning, pattern recognition, and robotics. Particularly, he is interested in the theories and algorithms that empower robot to see, think and conduct more like a human. Previously, he was awarded the national scholarship for graduate student, and listed in the first term of Huawei F(X) future scientist program member and the outstanding undergraduate of Southeast University. He has served as the reviewers of IEEE Computational Intelligence Magazine, Journal of Visual Communication and Image Representation and Journal of Electronic Imaging.
\end{IEEEbiography}
%
%
\begin{IEEEbiography}[{\includegraphics[width=1in,height=1.25in,clip,keepaspectratio]{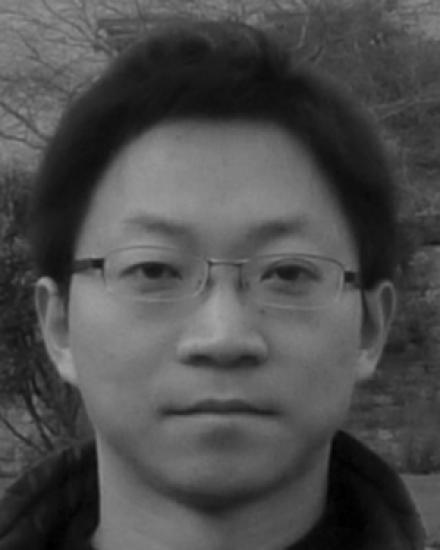}}]{Siyu Xia}
received his BE and MS degrees in automation engineering from Nanjing University of Aeronautics and Astronautics, Nanjing, China, in 2000 and 2003, respectively, and the PhD degree in pattern recognition and intelligence system from Southeast University, Nanjing, China, in 2006. He is currently working as an associate professor in the School of Automation at Southeast University, Nanjing, China. His research interests include object detection, applied machine learning, social media analysis, and intelligent vision systems. He was the recipient of the Science Research Famous Achievement Award in Higher Institution of China in 2015. He has served as the reviewer of many journals including TIP, T-SMC-B, T-IFS, T-MM, IJPRAI, and Neurocomputing. He received Outstanding Reviewer Award for Journal of Neurocomputing in 2016. He has also served on the PC/SPC for the conferences including AAAI, ACM-MM, ICME, and ICMLA. He is a member of IEEE and ACM.
\end{IEEEbiography}
\begin{IEEEbiography}[{\includegraphics[width=1in,height=1.25in,clip,keepaspectratio]{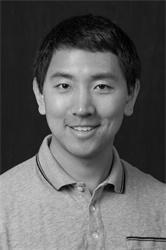}}]{Ming Shao} (S11-M16) received the B.E. degree in computer science, the B.S. degree in applied mathematics, and the M.E. degree in computer science from Beihang University, Beijing, China, in 2006, 2007, and 2010, respectively. He received the Ph.D. degree in computer engineering from Northeastern University, Boston MA, 2016. He is a tenure-track Assistant Professor affiliated with College of Engineering at the University of Massachusetts Dartmouth since 2016 Fall. His current research interests include sparse modeling, low-rank matrix analysis, deep learning, and applied machine learning on social media analytics. He was the recipient of the Presidential Fellowship of State University of New York at Buffalo from 2010 to 2012, and the best paper award winner/candidate of IEEE ICDM 2011 Workshop on Large Scale Visual Analytics, and ICME 2014. He has served as the reviewers for many IEEE Transactions journals including TPAMI, TKDE, TNNLS, TIP, and TMM. He has also served on the program committee for the conferences including AAAI, IJCAI, and FG. He is the Associate Editor of SPIE Journal of Electronic Imaging, and IEEE Computational Intelligence Magazine. He is a member of IEEE.
\end{IEEEbiography}
\begin{IEEEbiography}[{\includegraphics[width=1in,height=1.25in,clip,keepaspectratio]{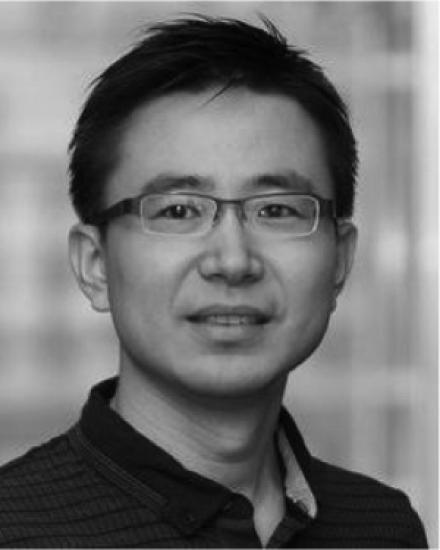}}]{Yun Fu}
(S’07-M’08-SM’11) received the B.Eng. degree in information engineering and the M.Eng. degree in pattern recognition and intelligence systems from Xi’an Jiaotong University, China, respectively, and the M.S. degree in statistics and the Ph.D. degree in electrical and computer engineering from the University of Illinois at Urbana-Champaign, respectively. He is an interdisciplinary faculty member affiliated with College of Engineering and the College of Computer and Information Science at Northeastern University since 2012. His research interests are Machine Learning, Computational Intelligence, Big Data Mining, Computer Vision, Pattern Recognition, and Cyber-Physical Systems. He has extensive publications in leading journals, books/book chapters and international conferences/workshops. He serves as associate editor, chairs, PC member and reviewer of many top journals and international conferences/workshops. He received seven Prestigious Young Investigator Awards from NAE, ONR, ARO, IEEE, INNS, UIUC, Grainger Foundation; nine Best Paper Awards from IEEE, IAPR, SPIE, SIAM; many major Industrial Research Awards from Google, Samsung, and Adobe, etc. He is currently an Associate Editor of the IEEE Transactions on Neural Networks and Leaning Systems (TNNLS). He is fellow of IEEE, IAPR, OSA and SPIE, a Lifetime Senior Member of ACM, Lifetime Member of AAAI, and Institute of Mathematical Statistics, member of Global Young Academy (GYA), INNS and Beckman Graduate Fellow during 2007-2008.
\end{IEEEbiography}




\end{document}